%% file: example_paper.tex
\documentclass{article}

\input{common}

\usepackage{microtype}
\usepackage{graphicx}
\usepackage{subfigure}
\usepackage{booktabs} % for professional tables

\usepackage{hyperref}

\usepackage[accepted]{icml2024}

\usepackage{amsmath}
\usepackage{amssymb}
\usepackage{mathtools}
\usepackage{amsthm}

\usepackage[capitalize,noabbrev]{cleveref}

\usepackage{units}
\usepackage{bbding}
\usepackage{diagbox}
\usepackage{bm}
\usepackage{makecell}
\usepackage{enumitem}
\usepackage{paralist}
\usepackage {setspace}
\usepackage{tablefootnote}

\theoremstyle{plain}
\newtheorem{theorem}{Theorem}[section]

\theoremstyle{definition}
\newtheorem{definition}[theorem]{Definition}

\theoremstyle{remark}

\usepackage[textsize=tiny]{todonotes}

\icmltitlerunning{Accelerating Transformer Pre-training with 2:4 Sparsity}

\begin{document}

\twocolumn[
\icmltitle{Accelerating Transformer Pre-training with 2:4 Sparsity}



\icmlsetsymbol{equal}{*}

\begin{icmlauthorlist}
\icmlauthor{Yuezhou Hu}{thu}
\icmlauthor{Kang Zhao}{}
\icmlauthor{Weiyu Huang}{thu}
\icmlauthor{Jianfei Chen}{thu}
\icmlauthor{Jun Zhu}{thu}
\end{icmlauthorlist}

\icmlaffiliation{thu}{Dept. of Comp. Sci. \& Tech., Institute for AI, BNRist Center, Tsinghua-Bosch Joint ML Center, THBI Lab, Tsinghua University}

\icmlcorrespondingauthor{Jianfei Chen}{jianfeic@tsinghua.edu.cn}

\icmlkeywords{Machine Learning, ICML}

\vskip 0.3in
]



\printAffiliationsAndNotice{}  % leave blank if no need to mention equal contribution

\begin{abstract}
Training large transformers is slow, but recent innovations on GPU architecture give us an advantage. NVIDIA Ampere GPUs can execute a fine-grained 2:4 sparse matrix multiplication twice as fast as its dense equivalent. In the light of this property, we comprehensively investigate the feasibility of accelerating feed-forward networks (FFNs) of transformers in pre-training. First, we define a ``flip rate'' to monitor the stability of a 2:4 training process. Utilizing this metric, we propose three techniques to preserve accuracy: to modify the sparse-refined straight-through estimator by applying the masked decay term on gradients, to determine a feasible decay factor in warm-up stage, and to enhance the model's quality by a dense fine-tuning procedure near the end of pre-training. Besides, we devise two techniques to practically accelerate training: to calculate transposable 2:4 masks by convolution, and to accelerate gated activation functions by reducing GPU L2 cache miss. Experiments show that our 2:4 sparse training algorithm achieves similar convergence to dense training algorithms on several transformer pre-training tasks, while  actual acceleration can be observed on different shapes of transformer block apparently. Our toolkit is available at \url{https://github.com/huyz2023/2by4-pretrain}.
\end{abstract}
\section{Introduction}
\label{submission}
Pre-training large-scale transformers is hard, for its intensive computation and time-consuming process~\cite{anthony2020carbontracker}. To accelerate training, sparsity-based methods have recently emerged as a promising solution, and one of the hardware-friendly sparse patterns is 2:4 sparsity. In a 2:4 sparse matrix, every four consecutive elements contain two zeros. Within a tensor core, a 2:4 sparse matrix multiplication (2:4-spMM) could be 2x faster than its dense equivalent on NVIDIA Ampere architecture GPUs.

Some works use 2:4 sparsity for accelerating training \cite{hubara2021accelerated, lu2023step, mcdanel2022accelerating, chmiel2023minimum}. However, they mainly target on convolutional neural networks (CNNs) \cite{hubara2021accelerated,mcdanel2022accelerating}, whose architecture, optimizer and training procedure are different from transformers.
% \jianfei{why transformers are different with cnns?}
Whether these 2:4 sparse training methods are capable for transformers remains under-explored. In practice, we find two barriers: 1) \textbf{Low accuracy.} The hyperparameters in some accuracy preserving techniques for transformers vary significantly from that for CNNs, which is ineffective if transplanted directly.
\emph{Remarkably, simply halving the inner dimensionality of a feed-forward network can also reduce the same amount of computational cost, but provides better performance than most of proposed 2:4 sparse training methods.}
2) \textbf{Inefficiency.} All previous works on 2:4 training stay on simulation, and do not provide actual acceleration results. Besides, they don't focus on other key operations beyond matrix multiplication that affect the practical time cost, such as overheads of pruning and activation functions. They usually lead to substantial mismatches between simulation and actual acceleration performance.

In this work, we aim to propose an end-to-end acceleration method for pre-training transformers based on 2:4 sparsity. Here are our major contributions:
\begin{compactitem}[$\bullet$]
\item We propose three accuracy-preserving techniques (two for masked decay and one for dense fine-tune) for 2:4 training. First, we propose to apply the masked decay on gradients rather than on weight. Second, we show that the feasible masked decay factor on transformers may be very small (100x smaller than it has been reported on CNNs) and devise a method to quickly determine an available decay factor. Besides, our analysis demonstrates that employing a dense fine-tuning stage at the end of pre-training, rather than at the beginning, can enhance the quality of transformers.
\item We analyze practical factors affecting the 2:4 training speed of transformers, which is rarely considered by previous works. We identify two speed bottlenecks: pruning overhead and gated activation functions' overhead. We proposed kernel-level accelerated methods to address each of these bottlenecks.
\item To the best of our knowledge, this is the first report on end-to-end acceleration  on pre-training transformers (\cref{acceleration-ffn}, \cref{whole-network}). Experiments show that transformers pre-trained using our proposed sparse training scheme are comparable or even superior in accuracy to those trained with dense training methods (Table \ref{BERT1}, \ref{gpt2}).
\end{compactitem}
\section{Related Work}
Existing sparsity-based methods can be classified into two categories: accelerating inference and accelerating training. For training acceleration, they can be further grouped by whether 2:4 sparsity is involved.

\paragraph{Sparsity for Inference Acceleration}
Early methods include one-shot pruning \cite{han2015learning, han2016deep, lee2018snip, mishra2021accelerating}. Later methods \cite{evci2021rigging,zhou2021learning,lasby2023dynamic} suggest using dynamic sparse training (DST). Particularly, \citet{zhou2021learning} proposes sparse-refined straight-through estimator (SR-STE) for 2:4 inference. Iterative magnitude-based pruning (IMP) methods \cite{chen2020lottery, chen2021earlyBERT, you2022drawing}, originated from the winning lottery ticket theory \cite{frankle2019lottery, frankle2020stabilizing}, can also be viewed as a DST 
 approach. All these methods only speedup the forward pass. They are insufficient to accelerate training.
 
\paragraph{2:4 Semi-Structured Sparsity for Training Acceleration}
Accelerating training by 2:4 sparsity is hard, because both the forward and backward passes need to be accelerated. On some GPUs involving sparse tensor cores, 2:4-spMMs perform 2x faster than dense GEMMs \cite{mishra2021accelerating, busatoexploiting}. In light of this, \cite{hubara2021accelerated} firstly proposes a transposable N:M mask to accelerate both output activations and input gradients computation in backward pass. \citet{zhang2023bidirectional} improve transposable mask to bi-directional mask (Bi-Mask) to further boost mask diversity. To accelerate calculating weight gradient via 2:4-spMM, an unbiased minimum-variance estimator (MVUE) is introduced \cite{chmiel2023minimum}. In addition, \citet{xu2022towards} also achieve fully sparse training of CNNs using spatial similarity. However, all these works do not report end-to-end training speedups on 2:4 sparse tensor cores, and they are built for CNNs. Practical 2:4 training acceleration on transformers has not been reported so far.

\paragraph{Other Structured Sparsity for Training Acceleration}
Structured sparsity means channel-wise pruning to dense networks. For instance, training a large model and then compressing it to be thinner or shallower seems effective \cite{li2020train, zhou2020go}, given a fixed accuracy requirement. However, it's not memory-efficient due to the larger model's redundancy. In addition, low-rank adaption proves to be an effective method to reduce fine-tuning costs \cite{hu2023llm}, but it can't accelerate the pre-training.

\section{Preliminary}
\label{preliminary}
In this section, we first present the mathematical formulations of dense training and fully sparse training. Afterward, we revisit the related methods which are helpful to achieve fully sparse training with 2:4 sparsity, including SR-STE \cite{zhou2021learning}, transposable N: M mask \cite{hubara2021accelerated}, and MVUE \cite{chmiel2023minimum}. 
\subsection{Dense Training}
\paragraph{Problem Formulation}
Dense training solves an optimization problem
$
\min _{\wv} \mathcal{L}(\wv)
$,
where $\mathcal{L}$ is a loss function, $\wv \in \mathbb{R}^D$ is the collection of dense weights of all layers, flattened to a vector. The loss is optimized by gradient descent optimization algorithms such as SGD, Adam \cite{kingma2017adam} and AdamW \cite{loshchilov2019decoupled}.
\paragraph{GEMMs of a Linear Layer in Dense Training}
In each training step, a single linear layer performs three general matrix multiplications (GEMMs):
\begin{align}\label{f:FWD}
\Zv=\Xv\Wv^{\top},~~~\nabla_{\Xv}=\nabla_{\Zv} \Wv,~~~\nabla_{\Wv}=\nabla_{\Zv}^\top  \Xv,
\end{align}
% \begin{equation}\label{f:FWD}
% \Zv=\Xv\Wv^{\top} \end{equation}
% \begin{equation}\label{f:BWD-Actv}
% \nabla_{\Xv}=\nabla_{\Zv} \Wv \end{equation}
% \begin{equation}\label{f:BWD-Weight}
% \nabla_{\Wv}=\nabla_{\Zv}^\top  \Xv
% \end{equation}
where $\Xv, \Wv$ and $\Zv$ are input activations, weights, and output activations, with shape $\Xv, \nabla_{\Xv} \in \mathbb{R}^{p \times q}$, 
$\Wv, \nabla_{\Wv} \in \mathbb{R}^{r \times q}$, 
and $\Zv, \nabla_{\Zv} \in \mathbb{R}^{p \times r}$. Here, the three GEMMs computes output activations, input activation gradients, and weight gradients, respectively. 
Without loss of generality, we assume the input $\Xv$ to be a 2D matrix rather than a 3D tensor. In the feed-forward networks of a transformer, this can be done by simply flattening the input tensors' first two axes, \ie, axes of batch size and sequence length.
\subsection{Fully Sparse Training with 2:4 Sparsity}\label{mvue-content}
GEMMs can be accelerated with structured sparsity. Particularly, 2:4 sparsity \cite{mishra2021accelerating} is a semi-structured sparsity pattern supported on NVIDIA Ampere architectures. A 2:4 sparse matrix partitions its elements into groups of four numbers, where each group has exactly two zeros. Depending on the direction of partition, there are row-wise 2:4 sparse matrix and column-wise 2:4 sparse matrix; see \cref{sec:24sparsity}. With such sparsity, a GEMM $\Cv=\Av\Bv$ can be accelerated by 2x with the 2:4-spMM kernel if either $\Av$ is  row-wise 2:4 sparse, or $\Bv$ is column-wise 2:4 sparse.

To accelerate training, each GEMM in \cref{f:FWD} should have one 2:4 sparse operand. In general, weights and output activation gradients are selected to be pruned due to relatively lower pruning-induced loss \cite{chmiel2023minimum}. That is,  
% check dims!
\begin{equation}\label{f:PFWD}
\Zv = \Xv S_{wt}( \Wv^{\top} ),
\end{equation}
\begin{equation}\label{f:PBWD-Actv}
\nabla_{\Xv} = \nabla_{\Zv} S_w(\Wv ),
\end{equation}
\begin{equation}\label{f:PBWD-Weight}
\nabla_{\Wv} = S_z(\nabla_{\Zv}^{\top} )  \Xv.
\end{equation}
In \cref{f:PFWD,f:PBWD-Actv,f:PBWD-Weight},  $S_{wt}, S_{w}$, and $S_{z}$ represent the pruning functions of $\Wv^{\top},\Wv$, and $\nabla_{\Zv}^\top$. They take dense matrices as input, and outputs 2:4 sparse matrices. By intuition, a pruning function picks out the 2 elements with the max magnitudes in the adjoining 4 elements and zero out the rest. With hardware support, computing \cref{f:PFWD,f:PBWD-Actv,f:PBWD-Weight} can be theoretically 2x faster than \cref{f:FWD}. This method use 2:4-spMMs for all matrix multiplications in forward and backward propagation, so we call it \emph{fully sparse training} (FST). Note that \cref{f:PBWD-Weight} contains a straight-through estimator (STE), which we will explain later.

% \jianfei{TODO: add some intuition that a pruning function simply takes the top 2 and zeros out other 2}
% On the other hand, \emph{dynamic sparse training} (DST) methods~\cite{???} only use spMM for forward propagation in Eq.~(\ref{f:PFWD}), and thus it can only accelerate inference by 2x. \jianfei{TODO: reconsider if we need to define DST here.}
\paragraph{Transposable Masks}
\citet{hubara2021accelerated} suggest that a weight matrix and its transpose can be simply pruned by multiplying binary masks, \ie, 
\begin{align*}
S_{wt}( \Wv^{\top} )=\Wv^{\top} \odot \Mv_{wt},~~~S_{w}( \Wv )=\Wv \odot \Mv_{w},
\end{align*}
where $\Mv_{wt}, \Mv_{w} \in \{0,1\}^{p \times q}$ are 2:4 sparse, and $\odot$ is element-wise product. To utilize 2:4-spMM, the two binary masks should be mutually transposable:   
\begin{equation}\label{f:transmask}
\Mv_{wt} = \Mv_{w}^{\top},
\end{equation}
which they call as transposable masks (same as our defination in \cref{transposalbemaskdefine}). In this manner, the backward pass share the same sparse weight matrix with the forward pass.
The authors also propose a 2-approximation method for generating such masks with claimed low computational complexity. 
\paragraph{Minimum-Variance Unbiased Estimator}
\citet{chmiel2023minimum} propose to calculate the 2:4 sparse masks of neural gradients by MVUE, \ie,
\begin{equation} \label{f:mvue}
    S_z(\nabla_{\Zv}^{\top} ) = \operatorname{MVUE}(\nabla_{\Zv}^{\top}).
\end{equation}
Compared to the commonly used minimum square error estimation, MVUE guarantees unbiasedness and minimizes the variance of the sparsified gradients, which is more favorable for promoting the convergence of training. 
\subsection{Optimization Strategies for Sparse Training}
The optimization of a sparse network is difficult as it has non-differentiable pruning functions. The optimization objective can be formulated as $\min _{\mathbf{w}} \mathcal{L}( \mathbf{\tilde w})$.
The network makes prediction with a sparse weight vector $\mathbf{\tilde w}=\mathbf{m}(\mathbf{w}) \odot \mathbf{w}$, where the mask $\mathbf{m}(\mathbf{w}) \in \left\{ 0,1 \right\}^{D}$ is the concatenation of masks for each layer. If a layer is not sparsified, then the corresponding mask is an all-one matrix. Computing the gradient is tricky since the mask $\mv$ is dynamically computed based on the dense weight $\wv$: by chain rule we have
$
\nabla_{\mathbf{w}} \mathcal{L}(\mathbf{\tilde w})=\frac{\partial \tilde \wv}{\partial \wv}\nabla_{\mathbf{\tilde w}} \mathcal{L}(\mathbf{\tilde w}),
$
% m 需要加粗吗？
where $\frac{\partial \tilde \wv}{\partial \wv}$ is a Jacobian matrix. However, $\tilde \wv$ is not differentiable with $\wv$ since it includes a non-differentiable mask-computing-function $\mv(\cdot)$ in it. Thus, it takes some skills to estimate the gradients and update the parameters.
\paragraph{STE} As $\tilde \wv$ is an approximation of $\wv$, a straight-through estimator (STE, \citet{bengio2013estimating}) directly passes the gradient of $\tilde \wv$ to $\wv$:
\begin{equation}
\nabla_{\mathbf{w}} \mathcal{L}(\mathbf{\tilde w}) \gets \nabla_{\mathbf{\tilde w}} \mathcal{L}(\mathbf{\tilde w}).
\end{equation}
\paragraph{SR-STE}
There is a problem with STE: only a portion of the weights in a layer participate in the forward calculation, but all the weights receive gradients. This indicates that the gradients associated with masked weights\footnote{Unlike some relevant literature, we use ``masked weights'' and ``pruned weights'' to denote the weights that are set to 0.} might be inaccurate. To suppress those inaccurate gradients, \citet{zhou2021learning} proposes sparse-refined straight-through estimator (SR-STE) which adds a decay term when updating:
\begin{align}
\label{eq:sgd}
\mathbf{w}_{t} \gets \mathbf{w}_{t-1}-\gamma(\nabla_{\mathbf{w}} \mathcal{L}_t({\mathbf{\tilde w}_{t-1}})
+ \lambda_W (\overline{\mathbf{m}(\mathbf{w}_{t-1})}) \odot \mathbf{w}_{t-1}),
\end{align}

where $\gamma$ stands for the learning rate, $\lambda_W$ is the decay factor, and $\overline{\mathbf{m}(\mathbf{w}_{t-1})}$ denotes the logical not operation of $\mathbf{m}(\mathbf{w}_{t-1})$. This decay term alleviates the change of weight mask. With SR-STE, the optimization target becomes
\begin{equation} \label{target}
\min _{\mathbf{w}} \mathcal{L} (\mathbf{\tilde w})+\tfrac{\lambda_W}{2} \Vert
\mathbf{w} \odot \overline{\mathbf{m}(\mathbf{w})}\Vert_2^2.
\end{equation}

\section{Accuracy Preserving Techniques}

While the methods reviewed in \cref{preliminary} can successfully perform FST on small-scale models such as ResNet and DenseNet, it is not clear whether they can be directly applied to pre-train large transformers. It is challenging for FST to preserve the accuracy of dense training, since the weights and masks need to be learned jointly, which is a non-differentiable, combinatorial optimization problem. Moreover, unlike inference acceleration methods, FST has no pre-trained dense model to start with. 
In this section, we propose three practical techniques to improve the convergence of FST for transformers: transformer-specific masked decay, Fast decay factor determination and dense fine-tuning.

\subsection{Flip Rate: Stability of Training } \label{stability}

Inspired by previous work \cite{zhou2021learning, you2022drawing}, we define a ``flip rate'' to measure how frequently the mask vector changes after one optimizer step. This metric could be used to monitor whether the network connection is stable during training.

\begin{figure}[!h]
    \centering
    \includegraphics[width=1\linewidth]{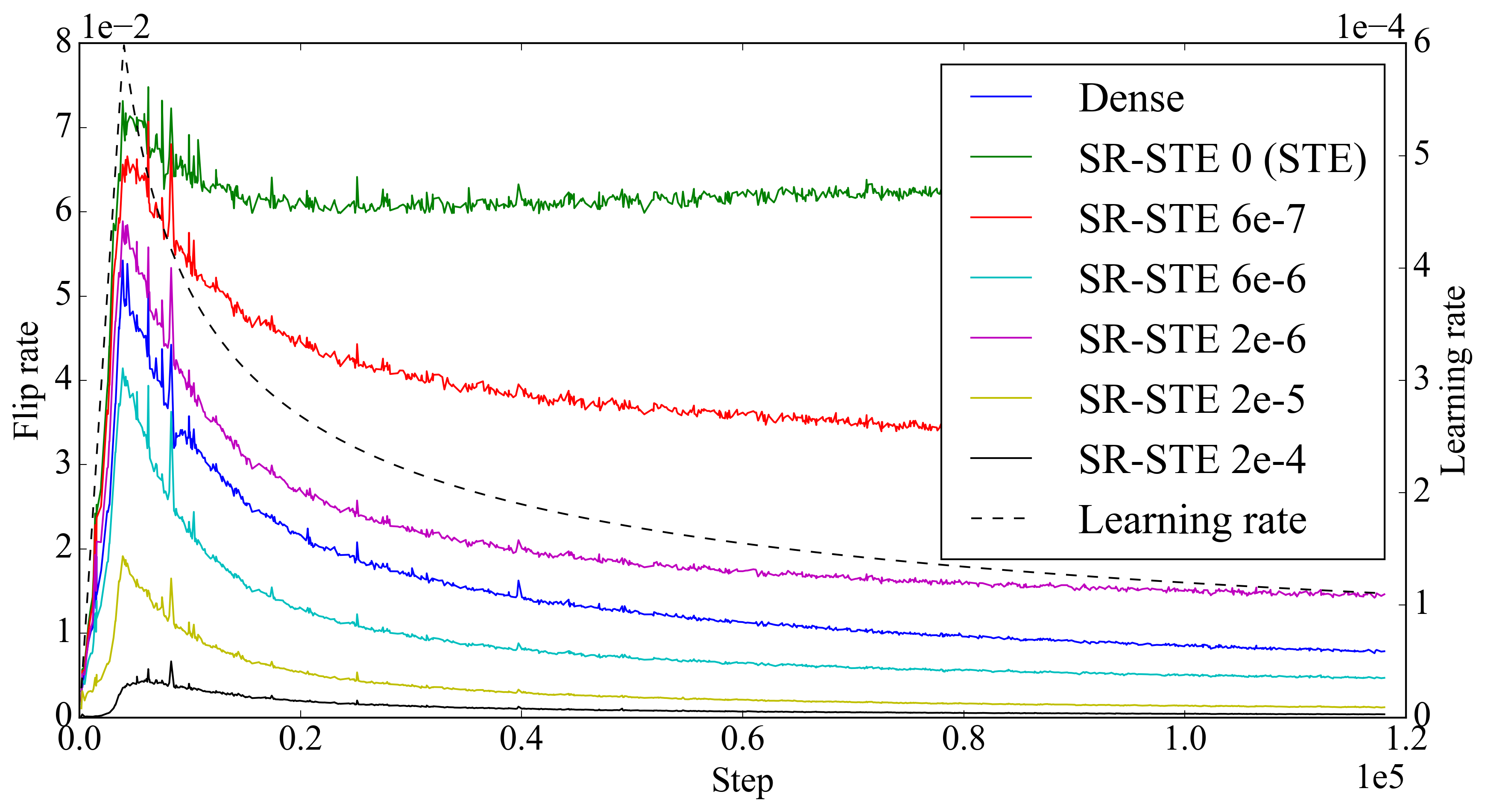}
    \caption{Flip rates change throughout the training of different $\lambda_W$ on Transformer-base. Note that these models utilize an identical learning rate schedule.}
    \label{fig:flip-rate}
\end{figure}

\begin{table}[!h]
\centering
\caption{Training results of different $\lambda_W$ on Transformer-base. As $\lambda_W$ increases from 0 to 2e-4, accuracy first rises and then drops, which means that $\lambda_W$ should be neither too big nor too small to reach the optimal results.}
\label{tab:lambda}
\vskip 0.15in
\begin{center}
\begin{small}
\begin{sc}
\begin{tabular}{llll}
\toprule
$\lambda_W$ & Avg epoch loss& Val loss & Test BLEU \\
\midrule
Dense       & 4.558                                                    &  3.978    &  26.15     \\
0 (STE)     &   4.76                                                  &   4.164   &  24.98    \\
6e-7        &    4.684                                                & 4.079     &  25.68    \\
6e-6        &    4.626                                                &   	4.033    & 25.81       \\
2e-6        &     	4.64                                                &  4.041    &  25.94     \\
2e-5        &       4.642                                              &  4.049     &  25.74    \\
2e-4        &        4.662                                             &  4.06    &    	25.62    \\
\bottomrule
\end{tabular}
\end{sc}
\end{small}
\end{center}
\vskip -0.1in
\end{table}

\begin{definition}
\label{def:flip-rate}
Suppose $\mathbf{w}_{t}$ is a $D$-dimensional weight vector at time $t$, and the flip rate $r_t$ is defined as the change in proportion of the mask vector after an optimizer step: $r_{t}=\Vert \mathbf{m}(\mathbf{w}_{t}) - \mathbf{m}(\mathbf{w}_{t-1}) \Vert_1/D \in [0, 1]$. The larger $r_t$ is, the more unstable the network connections become. 
\end{definition}

\citet{you2022drawing} suggest that a sparse neural network acts differently in different training phases. In the early phase of training, it eagerly explores different connection modes, which means the masks vector change rapidly over time. Later, the masks gradually become stable, and the network turns itself to fine-tune weight values. In terms of flip rate, we hypothesize that

\textit{A healthy training process comes with the flip rate $r_t$ rising at the beginning of training and then gradually fading to $0$.}

We measure flip rate change for dense training, STE and SR-STE with different $\lambda_W$ in \cref{fig:flip-rate}. For dense training, we compute the flip rate by pruning the dense weight in each iteration, despite the pruned weight is never used for training. In terms of flip rate, dense training is healthy: its $r_t$ exactly increases first before declines. If a training process consistently has higher flip rate than dense training, which we call as ``flip rate explosion'',  it may suffer from a loss in final accuracy due to unstable training; see \cref{tab:lambda}. In practice, STE suffers from a flip rate explosion, while SR-STE takes effect by ``freezing'' masks of weights: by adding a decay term, it decrease the number of flips. This inhibition effect is related to the decay factor of SR-STE: the larger $\lambda_W$ is, the stronger the inhibition of flips is, and the smaller flip rate goes.

In this section, all methods we propose involve our ultimate principle: \textit{the peak of the curve should be sufficiently high to fully explore different connection modes, and the tail should be sufficiently low for the optimization process to converge.}
% \jianfei{emphasize the goal: peak should be sufficiently high, tail should be sufficiently low}

% In contrast, STE demonstrates a consistently higher flip rate during training, which means the connections being unstable, and the method generally leads to deteriorating training loss. SR-STE lowers the flip rate by suppressing the mask change of pruned weights, therefore leading to improved training precision over naive STE. Moreover, the flip rates are closely related to decay factors: larger $\lambda_W$ results in smaller flip rate, which means the inhibition of $r_t$ is strengthened.
% \jianfei{TODO: more elaborate}

\subsection{Transformer-Specific Masked Decay }\label{masked-decay}
Based on our insights on flip rate, we propose a method to suppress the frequent change of masks during FST for transformers, which we call \emph{masked decay}. %In contrast to SR-STE, our proposed masked decay method has two features: the decay term is added to gradients and fast decay factor determination. 

% \paragraph{The Appropriate Placement of Mask Decay}
Unlike \cref{eq:sgd} which imposes regularization directly on weights, we propose to add masked decay on gradients, \ie,
\begin{align} \label{eq:adam-way}  
\mathbf{g}_t & \gets \nabla_{\mathbf{w}} \mathcal{L}_t(\mathbf{\tilde w}_{t-1}) 
+ \lambda_W (\overline{\mathbf{m}(\mathbf{w}_{t-1})} \odot \mathbf{w}_{t-1}).
\end{align}
On SGD, applying decay on weights and on gradients are equivalent, but on popular optimizers like Adam and AdamW they aren't. Specifically, Adam updates weights by
\begin{align}
\label{long}
    \wv_t \gets \wv_{t-1}-\frac{\gamma (\beta_1 \uv_{t-1}+(1-\beta_1)\gv_t)}{(1-\beta_1^t)(\sqrt{\hat{\vv}_t}+\epsilon)} 
\end{align}
where $\uv$ and $\vv$ are the first and second order momentum of $\wv$. Compared to \cref{eq:sgd}, the masked decay regularization term in \cref{eq:adam-way} would be later normalized by $\sqrt{\hat{\vv}_t}+\epsilon$ in \cref{long}, before it is subtracted from weights. In this way, each dimension receives a different intensity of decay (``masked decay''). More specifically, weights with larger gradients get smaller decay intensity, and vice versa.

\begin{figure}[!t]
    \centering
    \includegraphics[width=1\linewidth]{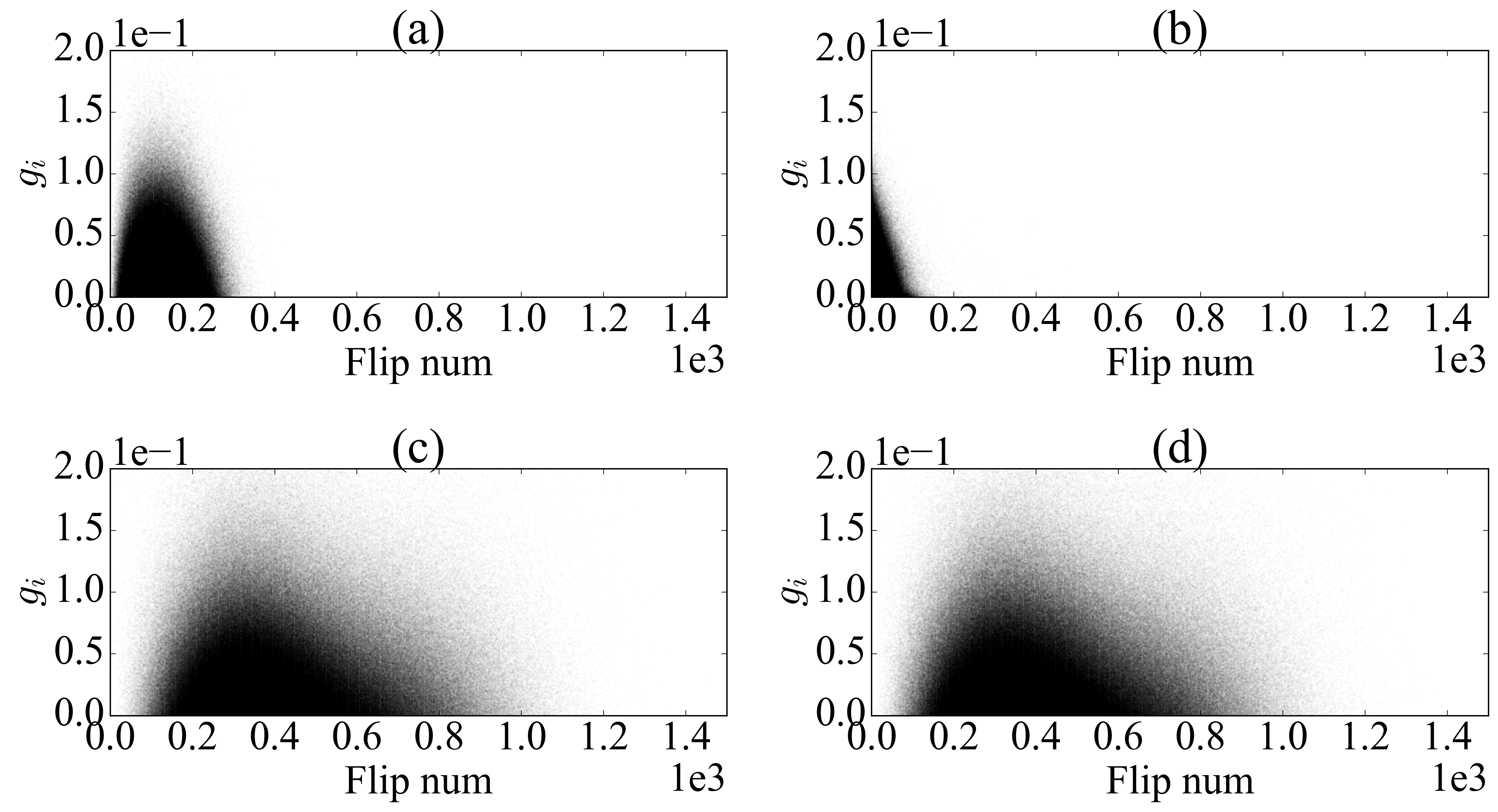}
    \caption{Scatter plots of cumulative flip number and L1 norm gap $g_i$ on every $4 \times 4$ block. All results are selected on Transformer-base, with epoch=20. (a) shows the result of dense model. (b)-(d) shows that of masked decaying on gradients, no decaying, and masked decaying on weights. Also, we do it on purpose to choose an extremely large $\lambda_W$ for SR-STE.}
    \label{fig:scatter}
\end{figure}

In FST, we periodically prune weights by their magnitudes. STE may cause the network to fall into such ``dilemma points'', where a portion of pruned weights and unpruned weights have nearly the same L1 norm. Thus, the network consistently oscillate between two possible masks $\mv_1$ and $\mv_2$, and is unlikely to jump out the dilemma itself. To illustrate this, we split each weight matrix by small $4 \times 4$ blocks. We count each block's cumulative flip number and measure the "L1 norm gap" by $ g_i = \norm{\mv_1 \odot \wv_i}_1 - \norm{\mv_2 \odot \wv_i}_1$, where $\wv_i$ is the $i$-th $4 \times 4$ weights, $\mv_1 \odot \wv_i$ and $\mv_2 \odot \wv_i$ have the first and second largest L1-norm among different pruning binary masks. The selected mask is most likely to oscillate between $\mathbf{m}_1$ and $\mathbf{m}_2$, especially when $g_i$ is small. In STE, there exists more $4 \times 4$ blocks with high flip num and low "L1 norm gap"; see \cref{fig:scatter}. This results in overall flip rate explosion of STE.

\begin{figure}[!t]
    \centering
    \includegraphics[width=1\linewidth]{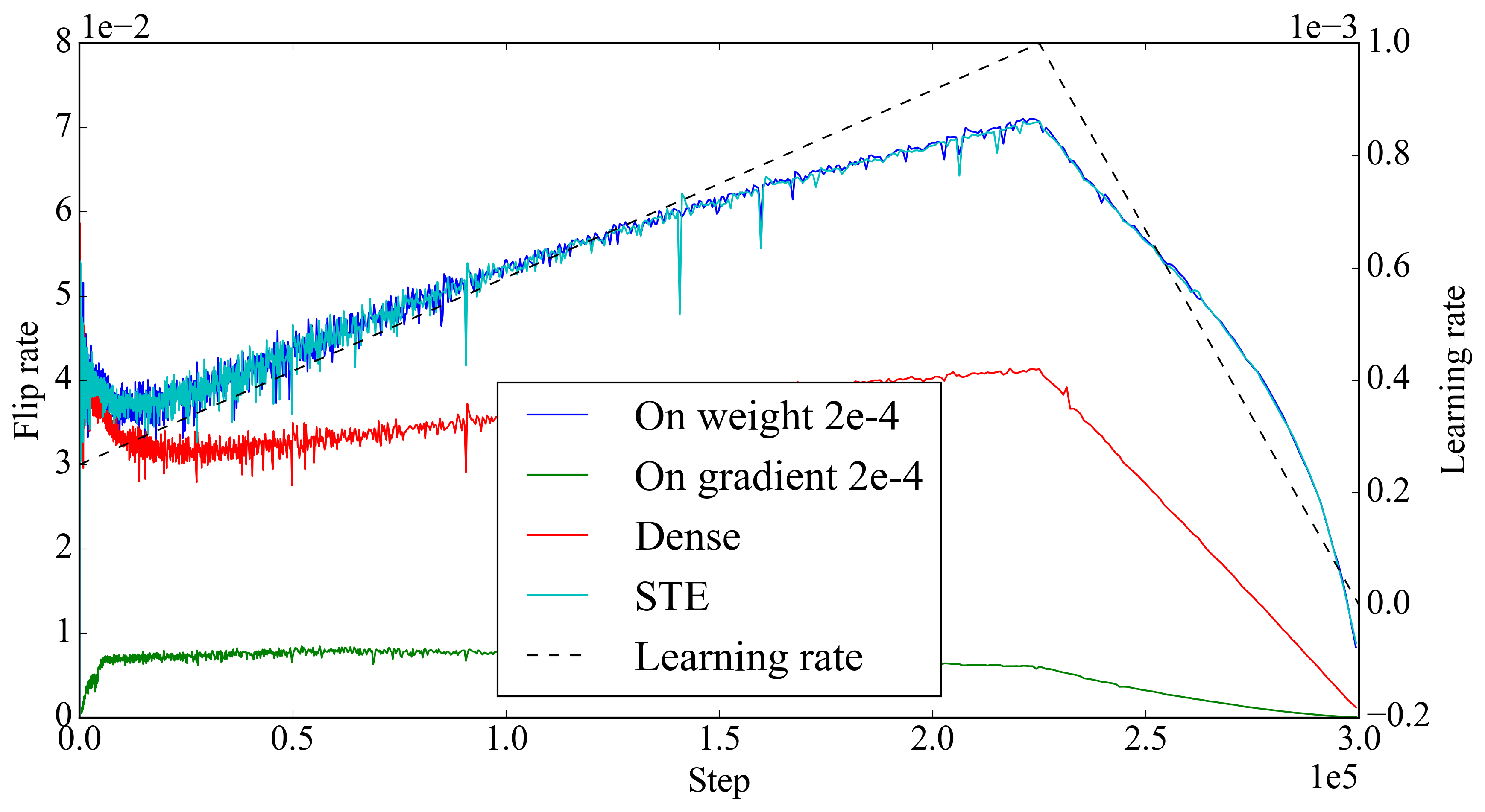}
    \caption{Applying masked decay on weights takes no effect to inhibit flip rate on BERT-base (compared to applying directly on gradient).}
    \label{fig:two-decay}
\end{figure}

\begin{table}[!t]
\centering
\caption{Optimal $\lambda_W$ for multiple models.}
\label{optimal-decay}
\vskip 0.15in
\begin{center}
\begin{small}
\begin{sc}
\begin{tabular}{lll}
\toprule
\multicolumn{2}{l}{Model}       & Optimal $\lambda_W$ \\
\midrule
\multicolumn{2}{l}{ResNet18 \cite{zhou2021learning}}    & 2e-4 \\
\multicolumn{2}{l}{BERT-base}        & 6e-6 \\
\multicolumn{2}{l}{Transformer-base} &  1e-6\\
\multicolumn{2}{l}{DeiT-tiny}        &  2e-3\\
\multirow{4}{*}{GPT-2}  & 124M  & 6e-5 \\
                        & 350M  & 2e-4 \\
                        & 774M  & 2e-4 \\
                        & 1558M & 6e-5 \\
                        \bottomrule
\end{tabular}
\end{sc}
\end{small}
\end{center}
\vskip -0.1in
\end{table}

On these occasions, we argue that an evenly masked decay  applied on weights is insufficient to save the training from such ``traps''. The weights don't differentiate themselves after an update, so masks may oscillate back.
By normalizing the weight gradients with $\sqrt{\hat{\vv}_t}+\epsilon$, our masked decay amplifies the regularization strength  for the dimension with smaller gradient, pushing it towards zero. Then, the regularized dimension can no longer compete with other dimensions. So we effectively break the tie and push the training process out of the trap, towards a ``healthier'' state.

The comparison results between our masked decay defined in \cref{eq:adam-way} and the conventional counterpart in \cref{eq:sgd} are shown in \cref{fig:two-decay}. Results show that applying masked decay on weights takes no effect to inhibit flip rate explosion of STE, while applying on gradients works fine.

\subsection{Fast Decay Factor Determination}
\label{fastdecaydetermine}
The determination of the decay factor $\lambda_W$ in \cref{eq:adam-way} is non-trivial: if $\lambda_W$ is excessively large, then the ``peak'' of the flip rate curve is not high enough; if $\lambda_W$ is too small, the ``tail'' of the curve is not low enough. Both do not provide a healthy training process.
% . On the one hand, in case $\lambda_W$ is excessively large, the mask decay-induced loss as shown in Eq.~(\ref{target}) would become the main focus of the optimizers, which probably leads to the competition with the regular cross-entropy loss and the increase of the final loss $\mathcal{L}( \mathbf{\tilde w})$. On the other hand, if $\lambda_W$ is too small, it may not effectively reduce the flip rate in the later phase of FST. \jianfei{maybe simply: larger lambda -> peak not high enough; smaller lambda -> tail not low enough}
Besides, we find that $\lambda_W$ values for CNNs and other small-scale networks differ significantly from those for transformers, while on transformers, optimal $\lambda_W$ can span up to three orders of magnitude (\cref{optimal-decay}).

% \jianfei{ok, now we convince the reader than tuning $\lambda$ is important. Why can't we simply grid search and observe the final accuracy then?}
As pre-training large transformers is costly,  grid searching for $\lambda_W$ with the final accuracy  is impractical, so it is vital to determine a feasible $\lambda_W$ as quickly as possible.
To quickly determine $\lambda_W$, here we propose a test-based method:

\begin{compactitem}[$\bullet$]
\item[1)] \textbf{Grid search on the warm-up stage of training.} For each $\lambda_W$ value in a candidate set, sample a corresponding flip rate of the sparse network from a small number of training steps. Note that sampling in early training stage is enough to obtain a representative flip rate specific to a sparse network.  
\item[2)] \textbf{Comparison with the dense counterparts.} Suppose $r_{t_0}$ to be the standard flip rate on the dense network at time $t_0$ and $r_{t_0}^{'}$ to be the sparse network's flip rate. Their ratio is $\mu =\nicefrac{r_{t_0}^{'}}{ \\r_{t_0}}$. We suggest that a feasible $\lambda_W$ should have $\mu \in [0.60, 0.95]$ and the sparse network may suffer from an accuracy drop if $\mu \geq 1$. 

\end{compactitem}

\subsection{Dense Fine-Tuning}\label{dense-finetuning}
\begin{figure}
    \centering
    \includegraphics[width=1\linewidth]{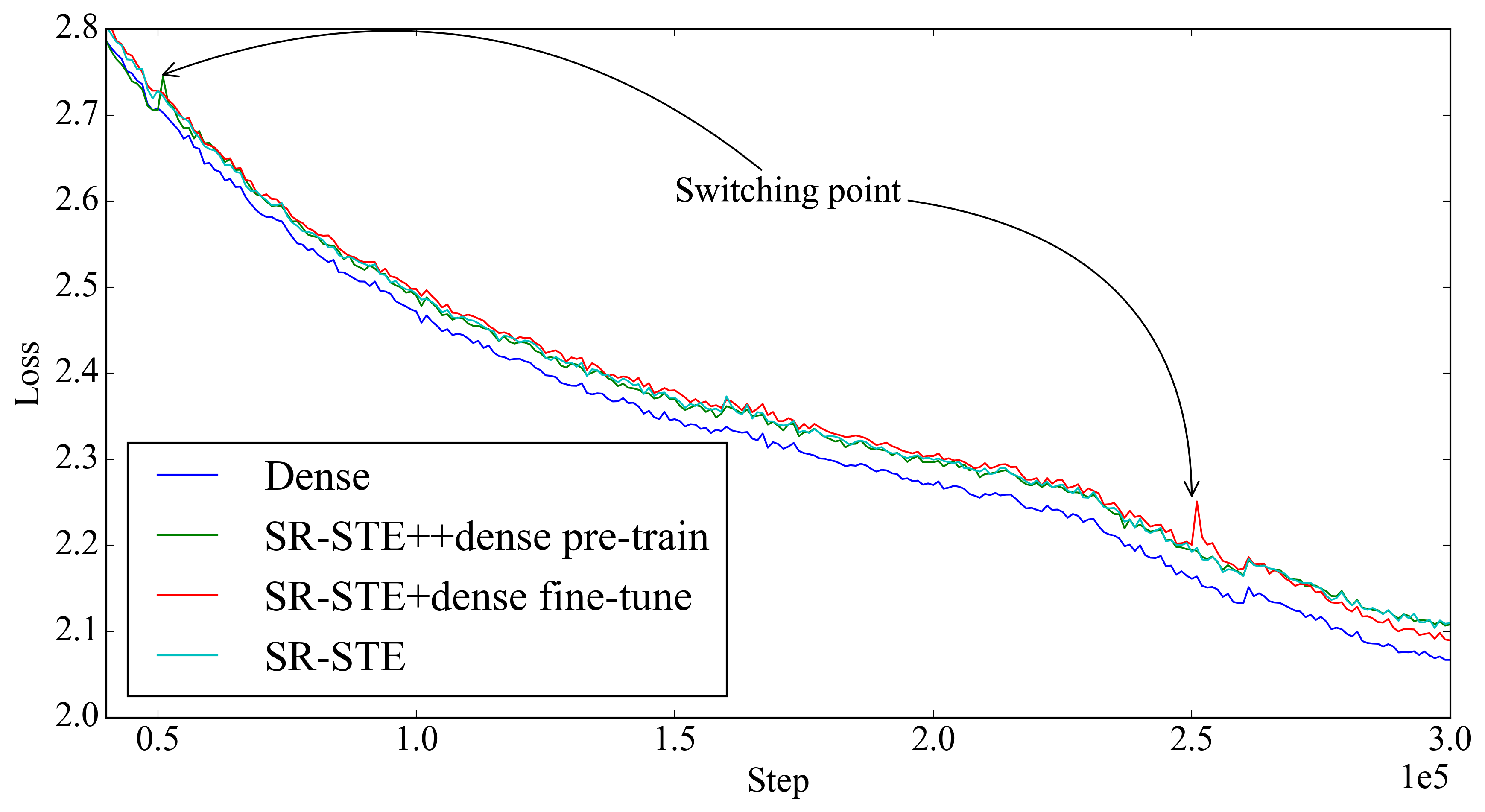}
    \caption{Dense fine-tuning versus dense pre-training on BERT-base}
    \label{fig:dense-finetune}
\end{figure}

To better improve accuracy, we suggest using a ``dense fine-tuning'' procedure at the end of training. Formally, we select a switch point $t_s$. FST is performed while $t \leq t_s$, and dense training is switched to if $t > t_s$.
\paragraph{Why Choose Dense Fine-Tuning Instead of Dense Pre-training?}
While previous work \cite{han2017dsd} suggest to switch between sparse and dense training stages, some recent works like STEP \cite{lu2023step} utilize dense pre-training rather than dense fine-tuning, which means a dense network is initially trained for a period of time before being switched to a sparse one. However, we argue that dense pre-training is meaningless in our FST process. As described in \cref{stability}, the peak of the flip rate curve should be sufficiently high to explore connection modes, so what matters most to the flip rate is the magnitudes of weights, which are the key to determine if connections are built or demolished. In this regard, both FST and dense pre-training are capable of delivering proper gradient magnitudes, so dense pre-training is a waste. The precise gradients are generally more necessary in the later stages of training, where the flip rate of the dense network comes to its tail. \cref{fig:dense-finetune} visualizes the loss curve of pre-training BERT-base, where dense pre-train obtains nearly the same result as the naive SR-STE method. From this, we propose the following insight:

\textit{
If dense pre-training of $t_\alpha$ steps provides slight improvement of accuracy, then moving the $t_\alpha$ dense steps to the end gives far more improvement than dense pre-training.
}

As for the specific position of the switch point in training, 
STEP \cite{lu2023step} suggests that the dense pre-training occupy $10\%$ to $50\%$ of the total steps. Likewise, we determine that our dense fine-tuning takes up the last $1/6$ of total steps for balance training efficiency and accuracy.

\section{Training Acceleration Techniques}\label{training acceleration techniques}

For transformers, the forward pass of FST involves pruning weights in FFNs with transposable 2:4 masks and then performing normal forward propagation. During backward propagation in FST, the gradients of input activations and weight gradients in FFNs are derived by \cref{f:PBWD-Actv} and (\ref{f:PBWD-Weight}), respectively. Note that we also utilize MVUE to prune gradients of output activations, \ie, \cref{f:mvue}.  
Compared to dense training, our FST replaces all the GEMMs in FFNs with 2:4-spMMs that theoretically perform 2x faster than their dense counterparts on GPUs within sparse tensor cores. 

In addition to speeding up the most time-consuming GEMMs in FFNs, there are three major operations that also have non-negligible impacts on training speed:

\begin{compactitem}[$\bullet$]
\item[1)] \textbf{Pruning.} In FST, pruning includes two steps: finding a mask that satisfies the 2:4 sparse patterns and then enforcing the mask to the corresponding dense matrices.  In our case, we find that the time cost of finding transposable masks is time-consuming.
\item[2)] \textbf{Activation functions.} In transformers, SwiGLU and GEGLU \cite{shazeer2020glu} are popular. These two activation functions involve a gate mechanism to regulate activations. This mechanism easily induces the GPU L2 cache misses, thus decreasing the computing speed.
\item[3)] \textbf{Updating optimizer states.} The excessive update frequency can introduce additional time overheads.
\end{compactitem}

Below, we show our methods to accelerate these operations, the main workflow of which is shown in \cref{sec:workflow}.

\subsection{Fast Computation of Transposable Masks}\label{transposalbemaskdefine}
\paragraph{Problem Formulation} We aim to find such a mask matrix $\Mv \in \{0,1\}^{r \times q}$ for every $\Wv \in \mathbb{R}^{r \times q}$ in the FFN layer that 1) each adjoining $4 \times 4$ block contains 8 non-zero positions; each row and column in the block occupies 2 non-zero elements exactly; 2) $\max _{\Mv} \Vert\Mv\odot\Wv\Vert_1$. Then $\Mv$ would be our targeting \textit{transposable mask}.
% \jianfei{you still don't describe the definition of the transposable mask.}

As described in \cref{f:transmask}, both a transposable mask itself and its transposition conform to the format of 2:4 sparsity. Previous 2-approximation algorithm \cite{hubara2021accelerated} consists of two steps: sort elements, and pick elements out of the array. They claim that the procedure has less computational complexity. However, in practice, the sorting and picking process contains too many jumps in its control flow, and may be fatal to modern GPU architecture.
To make full use of the GPUs' parallel computation capability (SIMD and SIMT), we convert the transposable mask-search process into a convolution operation which traverse all the masks to obtain the optimal one in three steps: 
\begin{compactitem}[$\bullet$]
\item[1)] Create a convolutional kernel in the shape of   $4 \times 4 \times n_t$, where  $n_t$ denotes the number of transposable masks. In the case of 2:4 sparsity, mask diversity $n_t = 90$. These mask blocks for 2:4 sparsity can be selected by exhaustively inspecting all potential masks offline.
\item[2)] Calculate the index matrix via \cref{alg:transposable-mask-search}. The index matrix denotes which $4\times 4$ mask in the convolutional kernel is the optimal mask that retains most of the weight norms after being applied to weights.
\begin{algorithm}[]
   \caption{transposable mask search}
   \label{alg:transposable-mask-search}
\begin{algorithmic}
   \STATE {\bfseries Input:} mask pattern $\mv'$, weight matrix $\Wv$
   \STATE 1. $\Wv = \operatorname{abs}(\Wv)$
   \STATE 2. $out = \operatorname{conv2d}(\Wv, \mv', stride=4, padding=0)$
   \STATE 3. $index = \operatorname{argmax}(out, dim=2)$
   \STATE \textbf{return} $index$
\end{algorithmic}
\end{algorithm}
\item[3)] Replace all the elements in the index matrix by the corresponding $4\times 4$ block, which is the desired mask.
\end{compactitem}

\begin{figure}[!ht]
    \centering
    \includegraphics[width=1\linewidth]{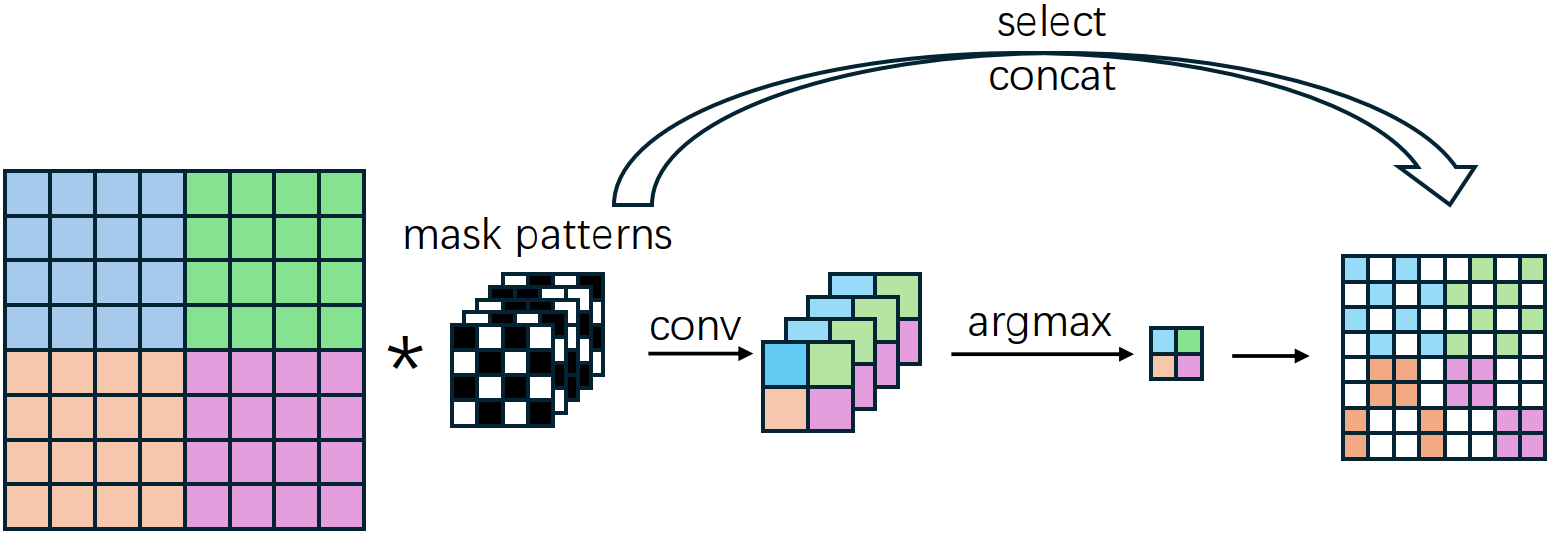}
    \caption{Transposable mask search}
    \label{fig:transposable-mask-search}
\end{figure}

\begin{figure}[!ht]
    \centering
    \includegraphics[width=1\linewidth]{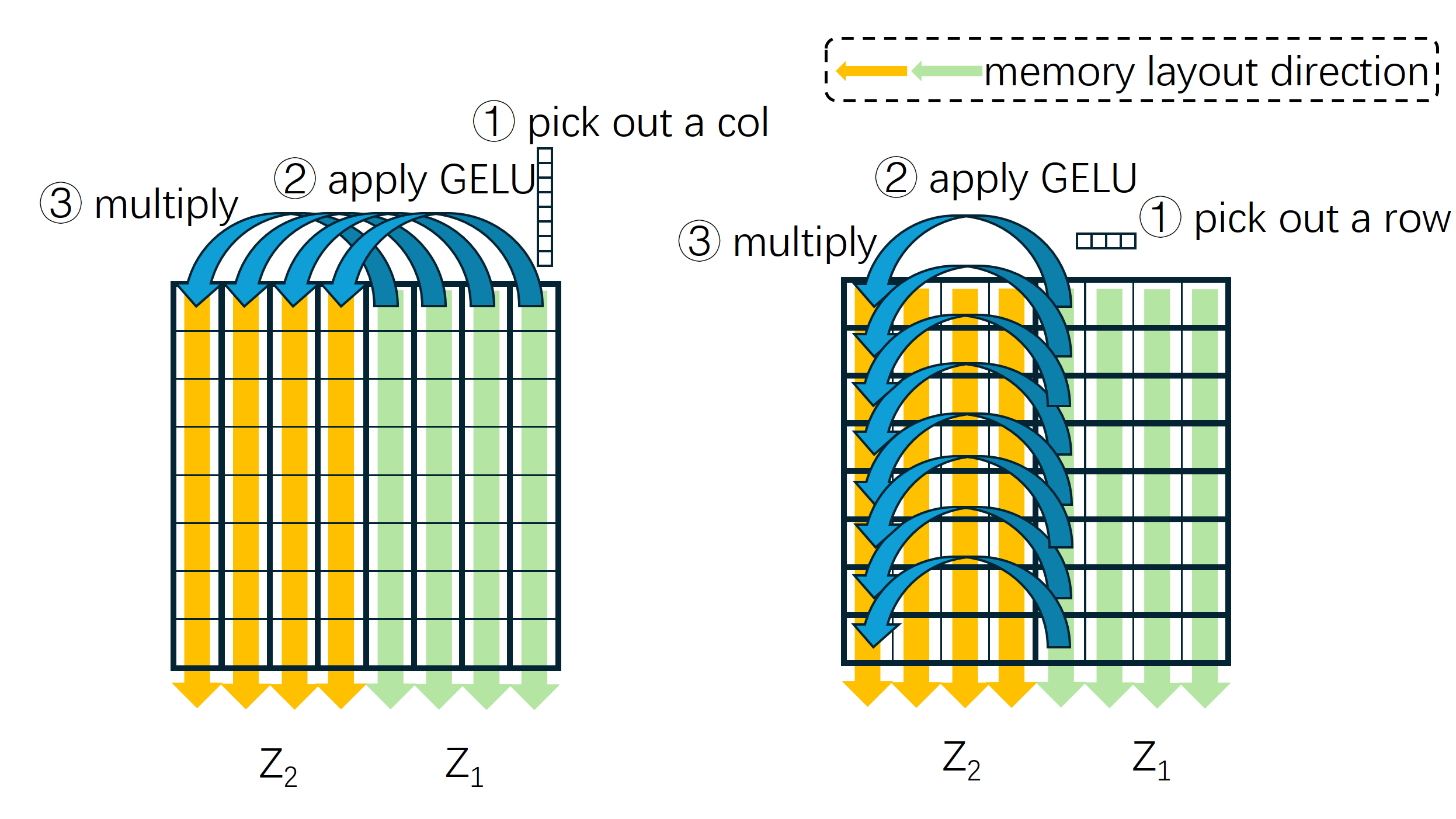}
    \caption{left: adapted method; right: intuitive method}
    \label{memory-layout-pic}
\end{figure}

\begin{table}[!h]
\centering
\caption{Throughput of two transposable search kernels on RTX3090 (TB/s).}
\label{throughput}
\vskip 0.15in
\begin{center}
\begin{small}
\begin{sc}
\begin{tabular}{lllll}
\toprule
\multirow{2}{*}{\diagbox{Input}{Method}} & \multicolumn{2}{l}{2-approx} & \multicolumn{2}{l}{Ours} \\
\cmidrule(lr){2-3} \cmidrule(lr){4-5}
                        & fp16          & fp32         & fp16        & fp32       \\
                        \midrule
$3072 \times 768$       & 18.5          & 36.4         & 69.2        & 104.7      \\
$4096 \times 1024$      & 22.5          & 38.4         & 91.9        & 131.5      \\
$5120 \times 1280$      & 22.6          & 44.4         & 91          & 128.2      \\
$1024 \times 1600$      & 22.8          & 44.8         & 95          & 134.5      \\
$8192 \times 2048$      & 23            & 45.1         & 99.4        & 142.9      \\
$16384 \times 4096$     & 23.2          & 45.4         & 100.1       & 144.8      \\
$30768 \times 8192$     & 23.2          & 45.5         & 100.9       & 145.1      \\
\bottomrule
\end{tabular}
\end{sc}
\end{small}
\end{center}
\vskip -0.1in
\end{table}

\begin{table}[!h]
\centering
\caption{Throughput of two GEGLU implementations on RTX3090 with fp16 column-major input tensors (TB/s).}
\label{throughput2}
\vskip 0.15in
\begin{center}
\begin{small}
\begin{sc}
\begin{tabular}{lllll}
\toprule
\diagbox{Input}{Method}  & Intuitive & Ours  \\
\midrule
$32 \times 512 \times 768$  & 18.4      & 55.5 \\
$32 \times 512 \times 1024$ & 19.9      & 55.7 \\
$32 \times 512 \times 1280$ & 18.2      & 55.9 \\
$32 \times 512 \times 1600$ & 18.4      & 55.9 \\
$32 \times 512 \times 2048$ & 19.5      & 56   \\
$32 \times 512 \times 4096$ & 11.8      & 56.1 \\
$32 \times 512 \times 8192$ & 12.1      & 56.2 \\
\bottomrule
\end{tabular}
\end{sc}
\end{small}
\end{center}
\vskip -0.1in
\end{table}

Notably, step (1) is executed offline. Step (2) and (3) are frequently performed during FST. The workflow of our method is shown in \cref{fig:transposable-mask-search}. Compared to the 2-approximation algorithm, our method is up to about 5 times faster (\cref{throughput}).

\subsection{ Acceleration of Gated Activation Functions  }

Activation functions with gated mechanisms are widely used in transformers such as GLM \cite{du2022glm} and LLaMA \cite{touvron2023llama}. Typical gated activation functions involve SwiGLU and GEGLU. The bottleneck of such activation functions is that the gate operations easily incur GPU L2 cache miss. Take GEGLU as an example: $\operatorname{GEGLU}(\Xv,\Uv,\Vv,\bv, \cv) = \operatorname{GELU}(\Xv \Uv^{\top} +\bv) \odot (\Xv\Vv^{\top} +\cv)$, where $\Xv \in \mathbb{R}^{p \times q},\Uv , \Vv \in \mathbb{R}^{r \times q},\bv ,\cv \in \mathbb{R}^{r}$.
In practice, this function is composed of three steps:
\begin{compactitem}[$\bullet$]
\item[1)] Concatenate $\Uv$ and $\Vv$ into a new weight matrix $\Wv \in \mathbb{R}^{2r \times q}$, and $\bv,\cv$ into a new bias vector $\dv \in \mathbb{R}^{2r}$.
\item[2)] Directly calculate $\Zv = \Xv \Wv^{\top}+\dv \in \mathbb{R}^{p \times 2r}$ as a compressed matrix.
\item[3)] Split the $\Zv$ in the second dimension into $\vect{Z_1},\vect{Z_2} \in \mathbb{R}^{p \times r}$. Calculate $\operatorname{GELU}(\vect{Z_1})\odot \vect{Z_2}$. 
\end{compactitem}

Different from dense model, where output activations are row-major matrices, in FST, the output activations are column-major; see \cref{appendixmemorylayout}. This property results in the third step being extremely time-consuming if conventionally $\Zv$ is accessed along the row dimension. To illustrate,  \cref{memory-layout-pic} shows that in a column-major matrix $\Zv$, accessing along the column accords with array layout. Thus, adjacent elements loaded into the GPU cache can be probably hit. By contrast, accessing along the row does not fully utilize the efficiency of GPU cache. In light of this, we carefully implement a GEGLU kernel where elements are accessed along the column dimension. In this way,  GEGLU is performed 5 times faster than the naive counterpart; see \cref{throughput2}.

\subsection{Other Implementation Details}
\paragraph{Reducing Updating Frequency}
We find that a 2:4 mask doesn't change a lot after one optimization step, and it is not necessary to update a mask frequently. For the sake of efficiency, we update the transposable masks of weights every $l$ optimizer steps. We usually take $l=40$ in practice.
\paragraph{Utilities}
For 2:4-spMMs, we use CUTLASS \cite{Thakkar_CUTLASS_2023}. Other GPU kernels are implemented in Triton, including transposable mask search kernel, pruning kernel, MVUE kernel, GEGLU kernel, and masked decay kernel.

\section{Experiments}

\begin{table*}[!h]
\centering
\caption{GLUE scores of different 2:4 training methods with BERT.}
\label{BERT1}
\vskip 0.15in
\begin{center}
\resizebox{\textwidth}{!}{
\begin{sc}
\begin{tabular}{llllllllllll}
\toprule
        Method      & Loss      & Avg score      & CoLA           & MNLI           & mnliextra        & MRPC & QNLI & QQP & RTE       & SST-2     & STS-B      \\
\midrule
Dense  & 2.0669    & $79.8 \pm 0.4$ & $45.3 \pm 1.1$ & $82.6 \pm 0.2$ & $83.4 \pm 0.1$   & $78.8 \pm 1.7 / 86.1 \pm 1$ & $89.3 \pm 0.2$ & $90.3 \pm 0.1 / 87.1 \pm 0$ & $55.8 \pm 0.9$ & $91 \pm 0.5$ & $83.7 \pm 1 / 83.7 \pm 1$ \\
Half  & 2.1280    & $77.9 \pm 0.4$ & $37.2 \pm 1.3$ & $82.4 \pm 0.1$ & $83 \pm 0.3$     & $75.1 \pm 1.4 / 84.2 \pm 0.7$ & $88.8 \pm 0.3$ & $89.9 \pm 0.1 / 86.6 \pm 0.1$ & $51.2 \pm 2.4$ & $92.1 \pm 0.5$ & $82.1 \pm 0.5 / 82.3 \pm 0.4$ \\
STEP   & 2.1179    & $77.7 \pm 0.1$ & $40.4 \pm 1.4$ & $82.2 \pm 0.1$ & $82.8 \pm 0.1$   & $74.5 \pm 0.7 / 83.5 \pm 0.4$ & $88.3 \pm 0.4$ & $90.2 \pm 0.1 / 87 \pm 0.1$ & $50.8 \pm 2.1$ & $92.3 \pm 0.3$ & $79.7 \pm 1.2 / 80.7 \pm 0.6$ \\
Bi-Mask & 2.1176    & $77.7 \pm 0.3$ & $38.3 \pm 0.7$ & $82.3 \pm 0.1$ & $83 \pm 0.1$     & $74.3 \pm 0.7 / 83 \pm 0.6$ & $88.3 \pm 0.3$ & $90.2 \pm 0.1 / 86.9 \pm 0.1$ & $53.1 \pm 1.4$ & $90.9 \pm 0.3$ & $80.9 \pm 0.7 / 81.7 \pm 0.4$ \\
\textbf{Ours} &\bm{$2.0968$     }&\bm{ $79.6 \pm 0.6$ }&\bm{ $44.4 \pm 1.9$ }&\bm{ $82.6 \pm 0.2$ }&\bm{ $83 \pm 0.1$     }&\bm{ $80.9 \pm 0.7 / 87.4 \pm 0.4$ }&\bm{ $88.4 \pm 0.3$ }&\bm{ $90.3 \pm 0.1 / 87 \pm 0.1$ }&\bm{ $54.3 \pm 1$ }&\bm{ $91.2 \pm 0.4$ }&\bm{ $82.9 \pm 2.1 / 83 \pm 1.7$ }\\
\bottomrule
\end{tabular}
\end{sc}
}
\end{center}
\vskip -0.1in
\end{table*}

\begin{table*}[!h]
\centering
\caption{GLUE scores with different model sizes on GPT-2 models.}
\label{gpt2}
\vskip 0.15in
\begin{center}
\resizebox{\textwidth}{!}{
\begin{sc}
\begin{tabular}{lllllllllllll}
\toprule
     Params & Method & Val loss & Avg Score  & CoLA & MNLI & MRPC & QNLI & QQP & RTE & SST-2 & STS-B & WNLI \\
     \midrule
\multirow{2}{*}{124M} & Dense  &2.907   & $73.9 \pm 1.1$ & $44.6 \pm 0.9$ & $82 \pm 0.1$ & $78.3 \pm 1.3/84.8 \pm 1$ & $88.4 \pm 0.2$ & $90 \pm 0$ & $86.5 \pm 0/61.3 \pm 1.5$& $91.9 \pm 0.2$ & $77.3 \pm 3.2/77.9 \pm 2.9$ & $24.3 \pm 7.1$\\
     & \textbf{Ours}&\bm{$2.952$    }&\bm{ $74.3 \pm 0.5$ }&\bm{ $44.8 \pm 1.3$ }&\bm{ $81.5 \pm 0.2$ }&\bm{ $77.5 \pm 1.8/84.2 \pm 1.3$ }&\bm{ $87.8 \pm 0.1$ }&\bm{ $89.5 \pm 0.1$ }&\bm{ $85.9 \pm 0.1/66 \pm 1$ }&\bm{ $90.6 \pm 0.4$ }&\bm{ $80 \pm 0.8/80.3 \pm 0.5$ }&\bm{ $23.9 \pm 6.4$}\\

\multirow{2}{*}{350M} & Dense& 2.618    & $76.3 \pm 0.1$                  & $54.3 \pm 0.4$ & $85.1 \pm 0.1$ & $80.7 \pm 1/86.6 \pm 0.7$ & $90.7 \pm 0.1$ & $91 \pm 0.1$ & $87.8 \pm 0.1/64.9 \pm 1.7$ & $93.5 \pm 0.4$ & $81.7 \pm 1.2/82.2 \pm 0.8$ & $17.6 \pm 3.2$\\
      &\textbf{Ours}&\bm{$2.688$}&\bm{$77.1 \pm 0.2$}&\bm{$51.8\pm1.8$}&\bm{$84.3\pm0.1$}&\bm{$80.6\pm1.3/86.5\pm0.8$}&\bm{ $90.4\pm0.2$}&\bm{$90.7\pm0.1$}&\bm{$87.5\pm0.1/66.7\pm1.3$}&\bm{$93.3\pm 0.4$}&\bm{ $83.4\pm1.1/83.5\pm1.1$}&\bm{$26.4\pm4$}\\
\multirow{2}{*}{774M} & Dense  &2.493   &   $76.2 \pm 0.4$                & $57.5 \pm 2$ & $86.1 \pm 0.1$ & $80.3 \pm 1.3$/$86.4 \pm 0.9$ & $91.4 \pm 0.2$ & $91.1 \pm 0.1$ & $88 \pm 0.1$/$67.7 \pm
 2.6$ & $94.6 \pm 0.4$ & $77.3 \pm 3.3$/$78.4 \pm 2.9$ & $15.1 \pm 2.3$\\
    &\textbf{Ours}&\bm{$2.564$}&\bm{   $77.1 \pm 0.4$                }&\bm{  $55.9 \pm 0.9$ }&\bm{ $85.6 \pm 0.2$ }&\bm{ $81.2 \pm 0.6/87 \pm 0.4$ }&\bm{ $91.4 \pm 0.1$ }&\bm{ $91 \pm 0.1$ }&\bm{ $87.8 \pm 0.1/71.5 \pm 0.7$ }&\bm{ $94.2 \pm 0.4$ }&\bm{ $81.8 \pm 1.3/82.3 \pm 1.2$ }&\bm{ $15.8 \pm 1.2$ }\\

\multirow{2}{*}{1558M} & Dense  &2.399   &   $76.5 \pm 0.5$ & $55.3 \pm 2$ & $87 \pm 0.1$ & $79 \pm 1/85.3 \pm 0.8$ & $91.8 \pm 0.3$ & $91.3 \pm 0.1$ & $88.3 \pm 0.1/73.3 \pm 2$ & $95.9 \pm 0.3$ & $78.5 \pm 2.4/79.2 \pm 2.5$ & $13 \pm 1.3$
\\
    &\textbf{Ours}&\bm{$2.489$ }&\bm{ $77.1 \pm 0.5$ }&\bm{ $56.4 \pm 3$ }&\bm{ $86.6 \pm 0.1$ }&\bm{ $80 \pm 0.4/86.1 \pm 0.3$ }&\bm{ $91.9 \pm 0.1$ }&\bm{ $91.4 \pm 0.1$ }&\bm{ $88.4 \pm 0.1/75 \pm 1.8$ }&\bm{ $95.2 \pm 0.4$ }&\bm{ $80.6 \pm 1.1/81.1 \pm 1.3$ }&\bm{ $12.7 \pm 1.1$} \\     
    \bottomrule
\end{tabular}
\end{sc}
}
\end{center}
\vskip -0.1in
\end{table*}

\begin{table}[!h]
\centering
\caption{SQuAD scores on GPT-2 models.}
\label{table:squad}
\vskip 0.15in
\begin{center}
\begin{small}
\begin{sc}
\begin{tabular}{llll}
\toprule
Params                & Method & EM   & F1   \\
\midrule
\multirow{2}{*}{124M} & Dense  & 67.6 & 78.8 \\
                      & \textbf{Ours}   & \bm{$67.5$} & \bm{$78.5$} \\
\multirow{2}{*}{350M} & Dense  & 73.2 & 83.6 \\
                      & \textbf{Ours}   & \bm{$71.9$} & \bm{$82.4$} \\
\multirow{2}{*}{774M} & Dense  & 74.3 & 84.9 \\
                      & \textbf{Ours}   & \bm{$74.3$} & \bm{$84.6$} \\
                      \bottomrule
\end{tabular}
\end{sc}
\end{small}
\end{center}
\vskip -0.1in
\end{table}

\begin{table}[!h]
\centering
\caption{Experimental results for DeiT.}
\label{table:deit}
\vskip 0.15in
\begin{center}
\begin{small}
\begin{sc}
\begin{tabular}{llll}
\toprule
Size    & Method & Acc@1   & Acc@5   \\
\midrule
\multirow{3}{*}{DeiT-tiny} & Original\tablefootnote{Results reported in the original paper; see \url{https://github.com/facebookresearch/deit/blob/main/README_deit.md}.}  & 72.2 & 91.1 \\
                      & Dense\tablefootnote{DeiT-base dense model using the original recipe.}  & 72.9 & 91.6 \\
                      & \textbf{Ours}   & \bm{$70.4$} & \bm{$90.1$} \\
\multirow{4}{*}{DeiT-small} & Original  & 79.9 & 90.5 \\
                      & Dense  & 79.9 & 94.5 \\
                      & Bi-Mask  & 77.6 & - \\
                      & \textbf{Ours}   & 79.2 & 94.8 \\
\multirow{3}{*}{DeiT-base} & Original  & 81.8 & 95.6 \\
                      & Dense  & 81.0 & 95.0 \\
                      & \textbf{Ours}   & \bm{$81.3$} & \bm{$95.4$} \\
                      \bottomrule
\end{tabular}
\end{sc}
\end{small}
\end{center}
\vskip -0.1in
\end{table}

\begin{table}[!h]
\centering
\caption{Experimental results for Transformer-base.}
\label{transformer}
\vskip 0.15in
\begin{center}
\begin{small}
\begin{sc}
\begin{tabular}{lllll}
\toprule
Method              & \makecell{Avg epoch\\loss} & \makecell{Test\\BLEU} & \makecell{Val\\BLEU} & Val loss  \\
              \midrule
Dense    &   4.558     &    26.15               &   26.56    &    3.982     \\
Half   &  4.659     &    26.12            &    26.36         &   4.041   \\
STEP          &  4.692       &    25.27              &   25.85       &   4.082     \\
\textbf{Ours}          &   \bm{$4.649$}     &     \bm{$26.48$}             &  \bm{$26.78$}       &   \bm{$3.977$}      \\
\bottomrule
\end{tabular}
\end{sc}
\end{small}
\end{center}
\vskip -0.1in
\end{table}

In this section, we validate the proposed training speedup methods on several transformers, including BERT \cite{devlin2019BERT}, GPT-2 \cite{Radford2019LanguageMA}, Transformer-base for machine translation \cite{vaswani2023attention}, and DeiT \cite{touvron2021training}. 
%this point is described before. For all networks, we replace the two \texttt{Linear} layers with our \texttt{TransposaableSparseLinear} layer in the Feed Forward block in every Encoder/Decoder block. We do not sparsify the linear projection in Attention blocks and the prediction head.
For BERT, we use Cramming \cite{geiping2022cramming} to pre-train a 16-layer BERT model with the sequence length of 512 on the C4 dataset \cite{2019t5}. 
%The trained BERT is evaluated on GLUE \cite{Wang2018GLUEAM}.
For GPT-2, we use nanoGPT \cite{nanogpt} to pre-train GPT-2 124M, 355M, 774M, and 1.5B on OpenWebText \cite{Gokaslan2019OpenWeb}. Both BERT and GPT-2 models are estimated on GLUE \cite{Wang2018GLUEAM}.
For DeiT \cite{pmlr-v139-touvron21a}, we pre-train DeiT-tiny on ImageNet-1K dataset \cite{5206848}.
Besides, we use fairseq \cite{ott2019fairseq} to train Transformer-base on the WMT 14 En-De dataset \cite{Bojar2014FindingsOT} and measure the BLEU \cite{article} score of the trained model. 

Of note, we use $n$ to denote the length of sequences, $d$ to denote the input and output dimensions of each transformer block, $d_{ff}$ to denote the inner dimensions of the FFNs in each transformer block, $h$ to denote the number of heads, and $N$ to denote the micro-batch size on each device. The pre-training and evaluation scripts are publicly available at \url{https://github.com/thu-ml/2by4-pretrain-acc-examples}.

\subsection{Accuracy Results}
To investigate the effect of different 2:4 sparse training methods,  we pre-train a sparse BERT-base model on the C4 dataset using two sparse training methods: STEP \cite{lu2023step} and Bi-Mask \cite{zhang2023bidirectional}. Besides, we also pre-train a dense BERT-base and a `Half' BERT-base for comparison. Of note, `Half' denotes a smaller yet still dense BERT-base model. To create  Half model, we simply reduce the $d_{ff}$ of each FFN layer in the original BERT-base by half while maintaining the original value of $d$. Theoretically, this adjustment halves the floating operations (FLOPs) of the original FFN layer as well. Except for the FFN layers, the shapes of the rest layers remain unaltered.

All the pre-trained models are measured on GLUE benchmark (WNLI excluded). Surprisingly, \cref{BERT1} shows that despite having identical FLOPs, the 2:4-sparse BERT-base trained with STEP and Bi-Mask shows inferior average scores compared to the Half model. The Half model attains an average score of 77.9 on GLUE tests, while STEP and Bi-Mask only reach 77.7 due to the weaknesses in MRPC, QNLI, and STSB. By comparison, BERT-base trained in our proposed training method achieves 79.6 on GLUE, which significantly outperforms other sparse training methods and is comparable with the dense baseline, \ie, 79.8.

\begin{table}[!t]
\centering
\caption{Experimental results of masked decay, MVUE, and dense fine-tuning (FT) with BERT-Base. For decay term, we use both techniques in \cref{masked-decay,fastdecaydetermine}.}
\label{ablation}
\vskip 0.15in
\begin{center}
\begin{small}
\begin{sc}
\begin{tabular}{lllll}
\toprule
\makecell{Masked\\decay}        & MVUE       & Dense FT  & Loss & Avg score \\
\bottomrule
\XSolidBrush &  \XSolidBrush          &    \XSolidBrush             & 2.1553     & $  77.6 \pm 0.2  $      \\
\Checkmark   &  \XSolidBrush          &     \XSolidBrush            & 2.1096     &   $79.2 \pm 0.2$        \\
\Checkmark   & \Checkmark &     \XSolidBrush            & 2.1172     &  $78.4 \pm 0.3  $       \\
\Checkmark   &    \XSolidBrush        & \Checkmark      & 2.0896     &  $79.4 \pm 0.2   $      \\
\Checkmark   & \Checkmark & \Checkmark      & \bm{$2.0968$}     &  \bm{$79.6 \pm 0.6$} \\
\bottomrule
\end{tabular}
\end{sc}
\end{small}
\end{center}
\vskip -0.1in
\end{table}

\begin{table}[!t]
\centering
\caption{Actual pre-train speed up on the whole network.}
\label{whole-network}
\vskip 0.15in
\begin{center}
\begin{small}
\begin{sc}
\begin{tabular}{lll}
\toprule
Parameters & Batch size & Speedup\\
\midrule
124M   & 16         & 1.18                \\
350M   & 8         & 1.2               \\
774M   & 4          & 1.21              \\
\bottomrule
\end{tabular}
\end{sc}
\end{small}
\end{center}
\vskip -0.1in
\end{table}

\begin{figure}[!t]
    \centering
    \includegraphics[width=1\linewidth]{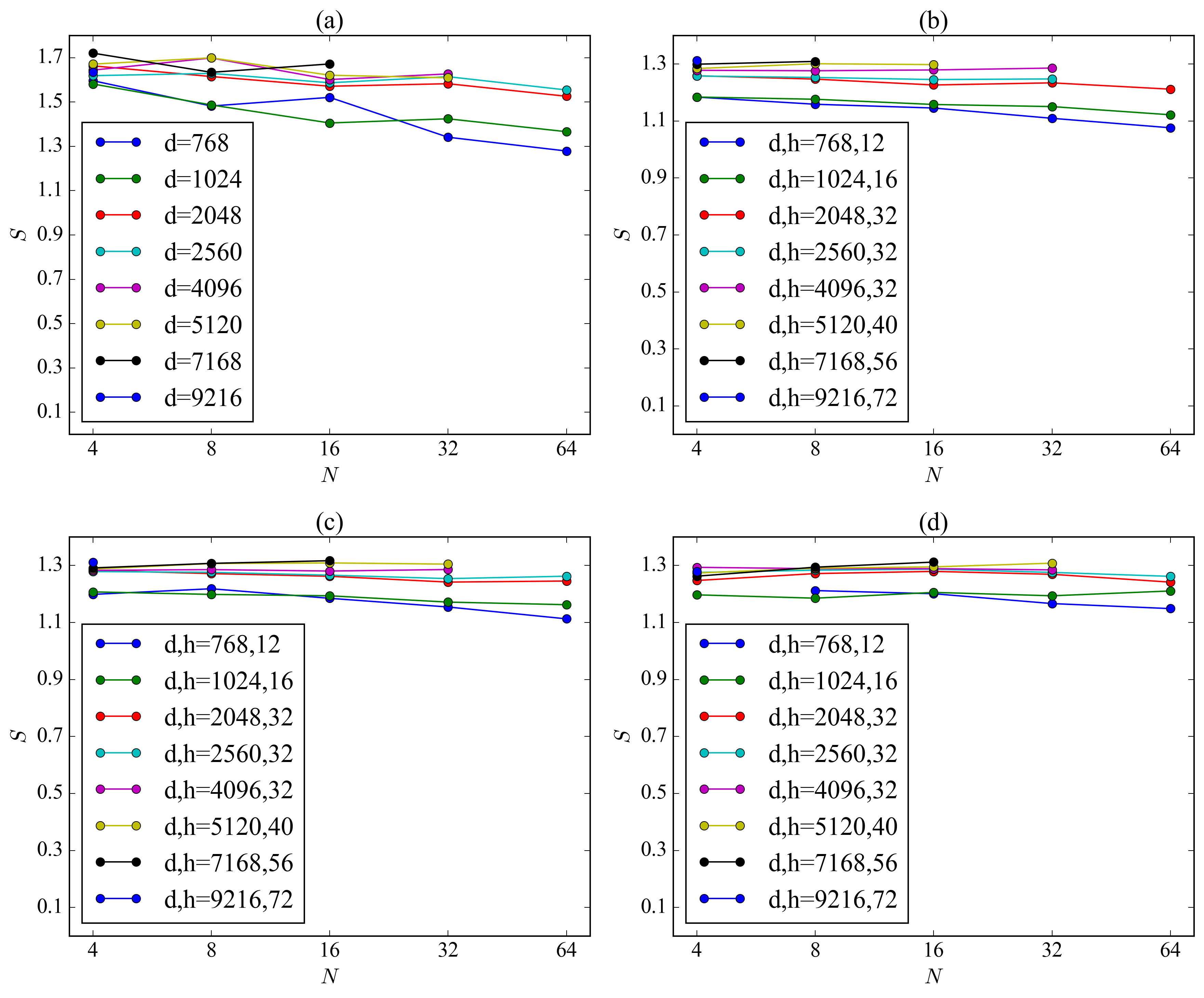}
    \caption{Result of acceleration ratio $S$ of different batch sizes and embedding Sizes. (a) shows the acceleration of a FFN layer. (b)-(d) shows the acceleration of a transformer block when $n=2048,1024,512$.}
    \label{acceleration-ffn}
\end{figure}

Besides, we pre-train GPT-2 models with proposed methods. Table \ref{gpt2} and \ref{table:squad} shows that our method for model sizes of 124M, 350M, 775M and 1558M achieves lossless scores compared with dense baselines. Similarly, DeiT and Transformer-base trained with our method also reach comparable results to dense training; see Table \ref{table:deit} and \ref{transformer}.
For GPT-2 and BERT, the training loss curves are sketched in \cref{sec:curve}.

\paragraph{Ablation Study}
We aim to investigate the effect of masked decay, MVUE and dense fine-tuning introduced in \cref{masked-decay}, \ref{mvue-content}, and \ref{dense-finetuning}. The  16-layer BERT-base is used for ablation study. 
Results in Table \ref{ablation} show that: 1) The dense fine-tuning procedure helps to improve accuracy on GLUE by 2 points at most ; 2) MVUE leads to insignificant, controllable accuracy loss; 3) By combining all these techniques together, 2:4 sparse training for transformers achieves comparable accuracy results as dense training.

\subsection{Speedup Results}
The training acceleration techniques proposed in \cref{training acceleration techniques} are evaluated using GPT-2 models and RTX3090 GPUs. FP16 mixed precision training is used on all models. The practical speedups of a single FFN layer, a single transformer block, and the entire network, compared to their respective dense counterparts, are reported. All the measured datum contain both forward and backward propagation.

\paragraph{Feed-forward Network Layers}~For a single FFN layer, we fix $n=2048$ and change $d$. Results in \cref{acceleration-ffn} show that a FFN layer can be accelerated up to 1.7x faster than its corresponding dense layer.

\paragraph{Transformer Block}
We measure the acceleration ratio of a transformer block when $n=512,1024,2048$. Results in \cref{acceleration-ffn} show that in most cases, a transformer block can be accelerated to 1.3x faster via 2:4 sparsity. To illustrate this, a detailed profile result is given in \cref{sec:profile}.

\paragraph{End-to-end Acceleration}
Finally, we test the practical speedups of training GPT-2 models. Results in \cref{whole-network} show that our training method conducts up to 1.2x faster than the dense training on a single RTX3090.

\section{Conclusions}
In this study, we are the first to propose accelerating the pre-training of transformers by 2:4 sparsity. We analyze the limitations of previous 2:4 training methods, including the impropriety in choosing positions and determining values of the masked decay factor, speed bottleneck incurred by computing transposable masks and gated activation functions. We propose a series of techniques to tackle them. Our training method is validated on DeiT, BERT, Transformer-base and GPT-2 models. In particular, we have attained 1.2x end-to-end training acceleration for the GPT-2 774M model without losing its accuracy.

\section*{Acknowledgements}
We would like to thank Ziteng Wang, Bingrui Li and Haocheng Xi for valuable discussions and help on the training large transformers. This work was supported by the National Key Research and Development Program of China (No.~2021ZD0110502), NSFC Projects (Nos.~62376131, 62061136001, 62106123, 62076147, U19A2081, 61972224), Tsinghua Institute for Guo Qiang, and the High Performance Computing Center, Tsinghua University. J.Z is
also supported by the XPlorer Prize.

\section*{Impact Statement}
Our proposed efficient algorithm can be used to accelerate pre-training large-scale transformers like GLM \cite{du2022glm}, LLaMA \cite{touvron2023llama}, etc. Recently, large transformers have exhibited remarkable efficacy in various fields such as natural language processing, computer vision, and speech recognition. However, the pre-training stage of large transformers is computationally intensive and time-consuming. For instance, pre-training a GPT-4 can span several months, even using a supercomputer equipped with thousands of GPUs. Thus, acceleration approaches are necessary. Our fully sparse training approach of transformers can potentially accelerate the FFN layers of a model by theoretical 2x faster, without loss of accuracy. Thus, it can be potentially used to save energy and reduce carbon footprint. But this work can also be used to accelerate baleful software, like software that generates malicious contents, which may have a negative impact on human society.

\nocite{langley00}

\bibliography{example_paper}
\bibliographystyle{icml2024}

%%%%%%%%%%%%%%%%%%%%%%%%%%%%%%%%%%%%%%%%%%%%%%%%%%%%%%%%%%%%%%%%%%%%%%%%%%%%%%%
%%%%%%%%%%%%%%%%%%%%%%%%%%%%%%%%%%%%%%%%%%%%%%%%%%%%%%%%%%%%%%%%%%%%%%%%%%%%%%%
% APPENDIX
%%%%%%%%%%%%%%%%%%%%%%%%%%%%%%%%%%%%%%%%%%%%%%%%%%%%%%%%%%%%%%%%%%%%%%%%%%%%%%%
%%%%%%%%%%%%%%%%%%%%%%%%%%%%%%%%%%%%%%%%%%%%%%%%%%%%%%%%%%%%%%%%%%%%%%%%%%%%%%%
\newpage
\appendix
\onecolumn
\section{2:4-spMM}
\subsection{2:4 Sparsity}\label{sec:24sparsity}
\begin{figure}[!h]
    \centering
    \includegraphics[width=0.5\linewidth]{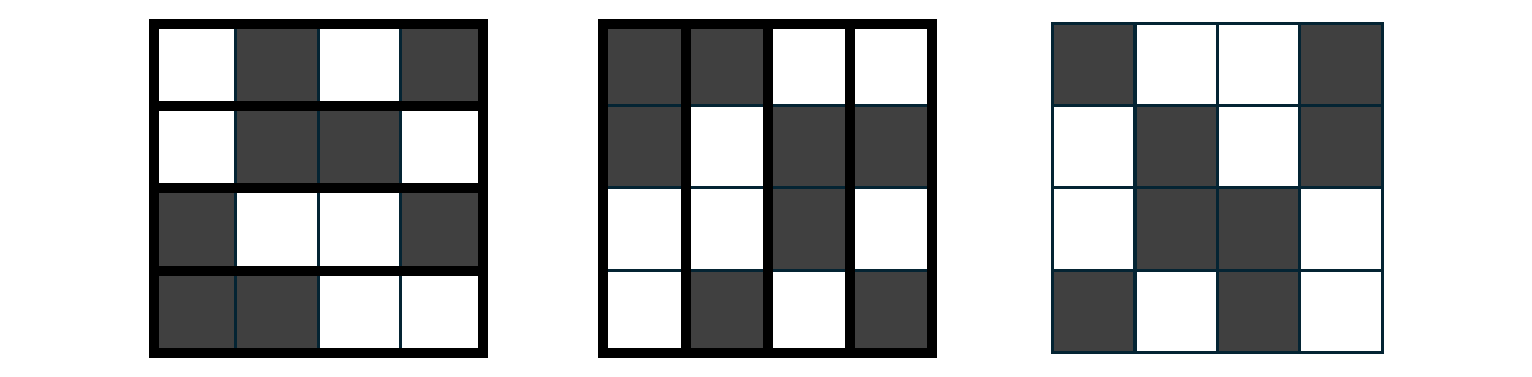}
    \caption{Row-wise 2:4, column-wise and transposable 2:4 sparse matrix.}
    \label{fig:2by4}
\end{figure}

Examples of row-wise, column-wise and transposable 2:4 sparse matrix are shown in \cref{fig:2by4}. Note that transposable 2:4 sparsity aligns with both row-wise and column-wise 2:4 sparsity.

\subsection{Array Layout}\label{appendixmemorylayout}
The array layout of different types of matrix multiplications are listed in \cref{memory-layout-table}, which explains why output activations and activation gradients are column-major matrices in FST.
\begin{table}[!h]
\centering
\caption{Array layout of $\Mv \Nv$. Here $S$ denotes that the matrix is in row-wise 2:4 sparsity, $R$ denotes row-major dense matrix, and $C$ denotes column-major dense matrix.}
\label{memory-layout-table}
\vskip 0.15in
\begin{center}
\begin{small}
\begin{sc}
\begin{tabular}{lllll}
\toprule
\diagbox{$\Mv$}{$\Nv$}   & $S$  & $S^\top$   & $R$ & $C$ \\
\midrule
$S$      & \XSolidBrush & \XSolidBrush      & $R$ & $R$    \\
$S^\top$   & \XSolidBrush & \XSolidBrush      & \XSolidBrush   & \XSolidBrush      \\
$R$    & \XSolidBrush & $C$ & $R$ & $R$    \\
$C$ & \XSolidBrush & $C$ & $R$ & $R$  \\
\bottomrule
\end{tabular}
\end{sc}
\end{small}
\end{center}
\vskip -0.1in
\end{table}

\section{Workflow}\label{sec:workflow}
\begin{figure*}[!h]
% \label{workflow}
    \centering
    \includegraphics[width=1\linewidth]{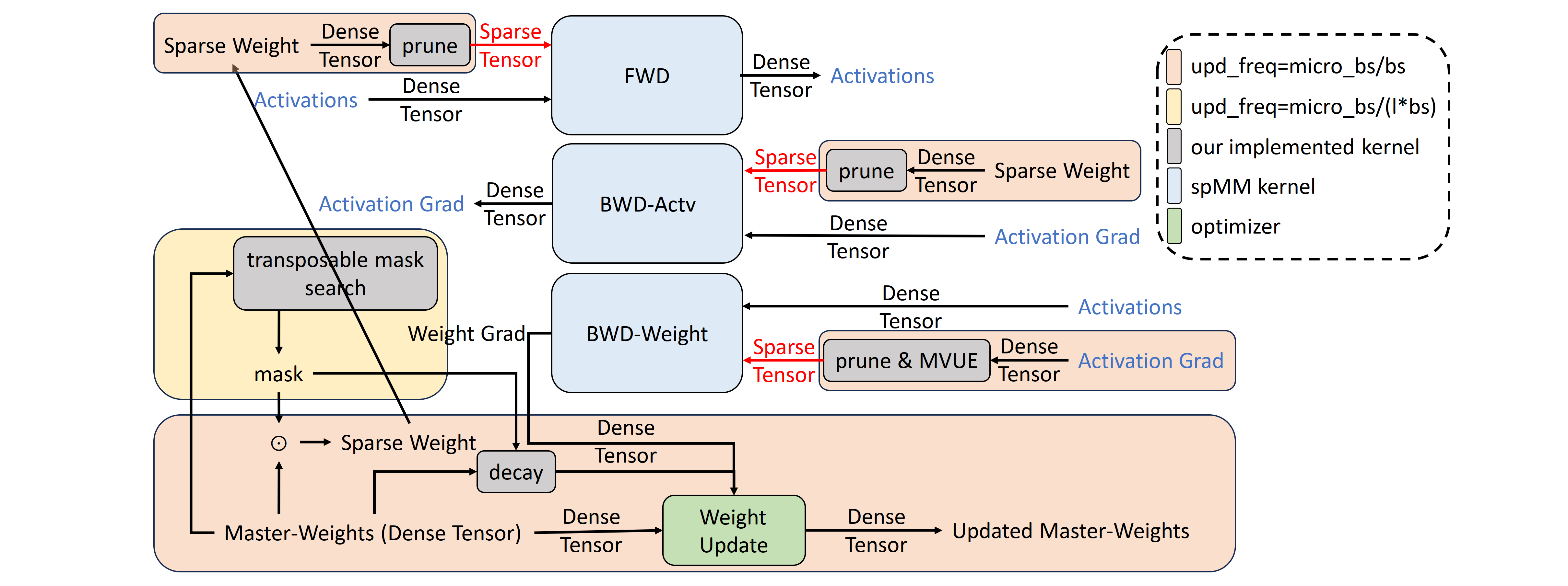}
    \caption{2:4 sparse training iteration for a layer on a single batch.}
    \label{workflow}
\end{figure*}

% \subsection{Forward and Backward Pass of a Linear Layer}
The main workflow of a single linear layer in FST process is depicted in \cref{workflow}.

% \subsection{Complete Workflow}
% The complete workflow of the training procedure is formulated in Algorithm \cref{alg:complete}.
% \begin{algorithm}[]
%    \caption{complete workflow}
%    \label{alg:complete}
%    \begin{algorithmic}
%    \STATE {\bfseries Input:} data $x_i$, weight $W$, total steps $t$, gradient accumulation steps $n$, mask update interval $l$
%    \REPEAT
%    \STATE Initialize $i=0$.
%    \STATE Initialize $m=\operatorname{transposable\_mask}(W)$.
%    \STATE Initialize $sparse\_W = \operatorname{prune}(W \odot m)$.
%    \STATE Initialize $sparse\_W\_T = \operatorname{prune}(W^T \odot m^T)$.
%    \STATE {\bfseries FWD:} $z = x_i \cdot sparse\_W^T$
%    \STATE {\bfseries BWD-Actv:} $\nabla x = z \cdot sparse\_W\_T^T$
%    \STATE {\bfseries BWD-Weight:} $\nabla W = \nabla W + \operatorname{MVUEprune}(z) \cdot x_i$
%    \IF{$i \mod n \cdot l$ is $0$}
%    \STATE $m=\operatorname{transposable\_mask}(W)$
%    \ENDIF
%    \IF{$i \mod n$ is $0$}
%    \STATE $\nabla W = \nabla W + \bm{\lambda (\sim m) \odot W}$
%    \STATE call optimizer
%    \STATE $sparse\_W = \operatorname{prune}(W \odot m)$
%    \STATE $sparse\_W\_T = \operatorname{prune}(W^T \odot m^T)$
%    \ENDIF
%    \STATE $i = i+1$
%    \UNTIL{$i \geq 5t/6$}
%    \REPEAT
%    \STATE {\bfseries FWD:} $z = x \cdot W^T$
%    \STATE {\bfseries BWD-Actv:} $\nabla x = z \cdot W$
%    \STATE {\bfseries BWD-Weight:} $\nabla W = z \cdot x$
%    \IF{$i \mod n$ is $0$}
%    \STATE call optimizer
%    \ENDIF
%    \STATE $i = i+1$
%    \UNTIL{$i \geq t$}   
% \end{algorithmic}
% \end{algorithm}

\section{Training Loss Curve}\label{sec:curve}
\begin{figure*}[h]
  \centering
  \includegraphics[scale=0.4]{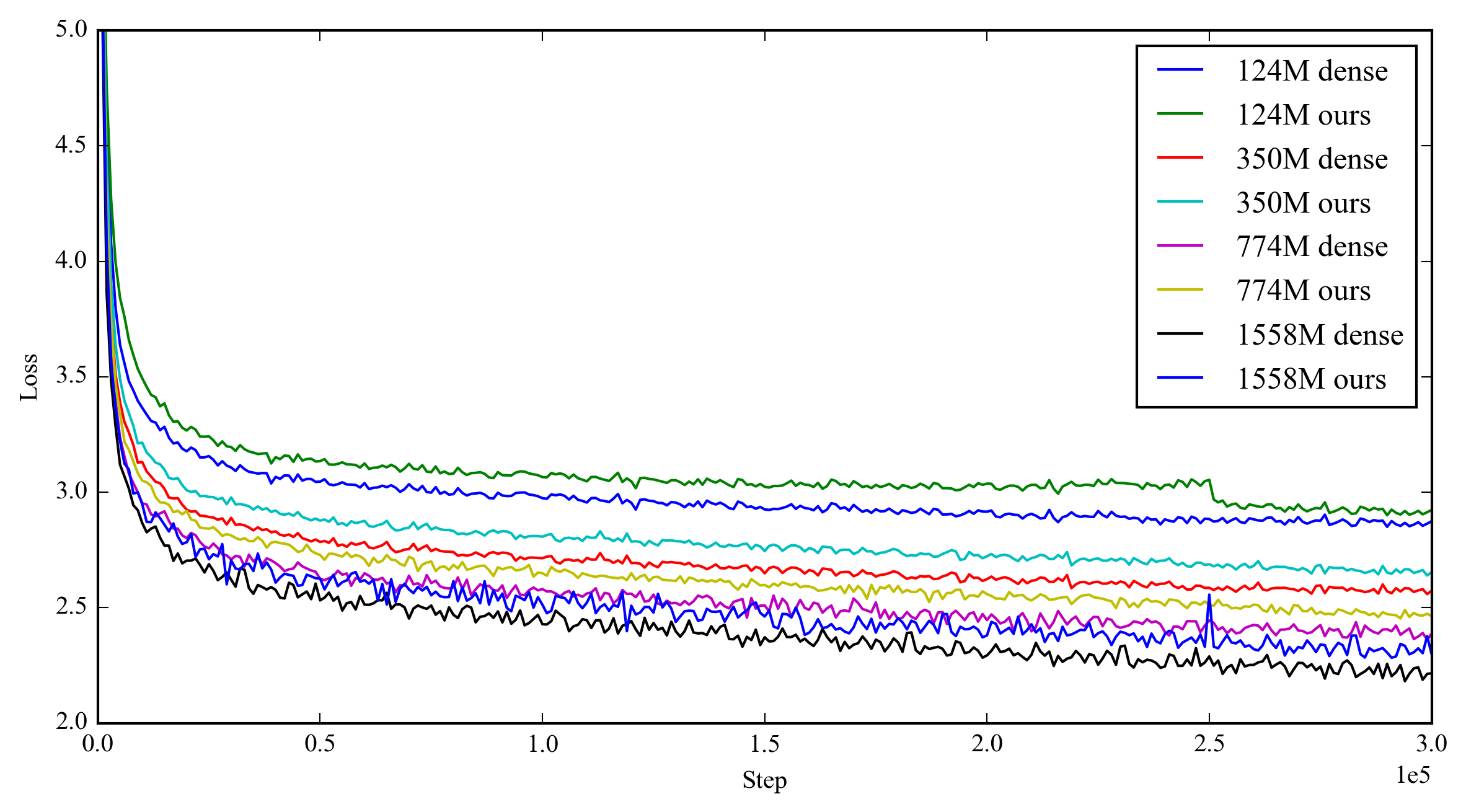}
  \hspace{0.5cm}
  \includegraphics[scale=0.4]{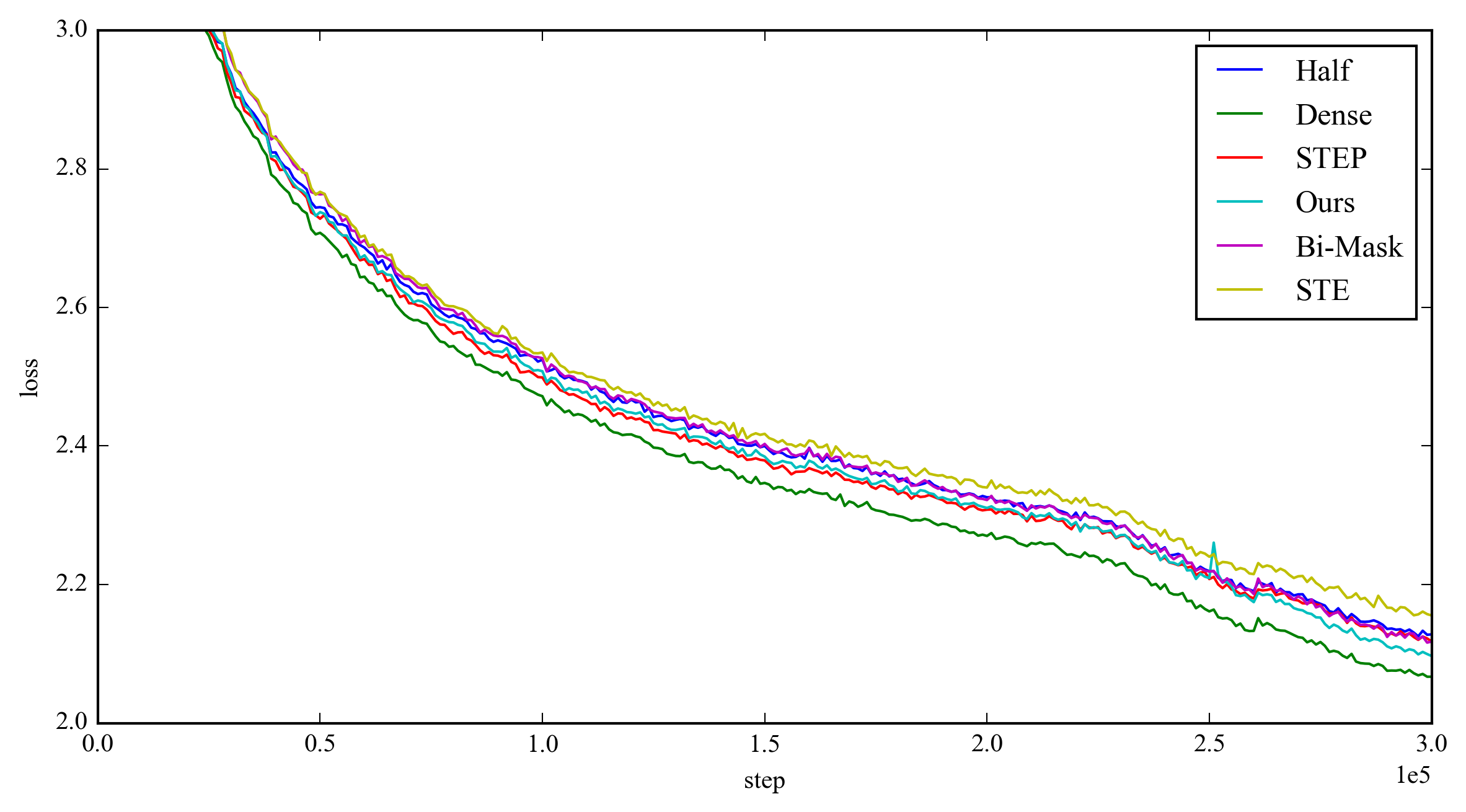}
  \caption{Left: train loss of GPT-2; right: train loss of BERT.}
  \label{loss-curve}
\end{figure*}

For BERT-base and GPT-2, we depict training loss curve in \cref{loss-curve}.

\section{Profiling result}\label{sec:profile}
To explain how we reach 1.3x block speedup, we profile our code and break down the time costs as shown in the table below; see Table \ref{table:prof}.

\begin{table}[h]
\centering
\caption{Time costs of each part of our network and the dense model in one iteration per layer. $m$ denotes the accumulation steps over micro batches. Our method is evaluated on GPT-2, with batch size 16, sequence length 1024, embedding dimension 1024 and heads number 16.}
\label{table:prof}
\vskip 0.15in
\begin{center}
\begin{small}
\begin{sc}
\begin{tabular}{|cccc|c|c|c|c|}
\hline
\multicolumn{1}{|c|}{}                      & \multicolumn{1}{c|}{}                        & \multicolumn{1}{c|}{}                     &            & Dense (ms/exec)  & Sparse (ms/exec) & \makecell{Acceleration\\ratio $S$}                    & \makecell{Frequency\\(exec/iter)}     \\ \hline
\multicolumn{1}{|c|}{\multirow{11}{*}{FFN}} & \multicolumn{1}{c|}{\multirow{5}{*}{Linear}} & \multicolumn{1}{c|}{FWD}                  & GEMM       & 12173.8          & 7305.78          & 1.666324472          & -        \\ \cline{3-8} 
\multicolumn{1}{|c|}{}                      & \multicolumn{1}{c|}{}                        & \multicolumn{1}{c|}{\multirow{3}{*}{BWD}} & GEMM       & 23295            & 14080.82         & 1.654378083          & -        \\ \cline{4-8} 
\multicolumn{1}{|c|}{}                      & \multicolumn{1}{c|}{}                        & \multicolumn{1}{c|}{}                     & MVUE+prune & 0                & 171.4            & -                    & -        \\ \cline{4-8} 
\multicolumn{1}{|c|}{}                      & \multicolumn{1}{c|}{}                        & \multicolumn{1}{c|}{}                     & Total      & 23295            & 14252.22         & 1.634482207          & -        \\ \cline{3-8} 
\multicolumn{1}{|c|}{}                      & \multicolumn{1}{c|}{}                        & \multicolumn{2}{c|}{\textbf{Total}}                    & \textbf{35468.8} & \textbf{21558}   & \textbf{1.645273216} & -        \\ \cline{2-8} 
\multicolumn{1}{|c|}{}                      & \multicolumn{1}{c|}{\multirow{3}{*}{Others\tablefootnote{All functions in FFN except linear layers, \ie, activation function and dropout.}}} & \multicolumn{2}{c|}{FWD}                               & 167              & 118.17           & -                    & -        \\ \cline{3-8} 
\multicolumn{1}{|c|}{}                      & \multicolumn{1}{c|}{}                        & \multicolumn{2}{c|}{BWD}                               & 65.5             & 20.03            & -                    & -        \\ \cline{3-8} 
\multicolumn{1}{|c|}{}                      & \multicolumn{1}{c|}{}                        & \multicolumn{2}{c|}{Total}                             & 232.5            & 138.2            & -                    & -        \\ \cline{2-8} 
\multicolumn{1}{|c|}{}                      & \multicolumn{1}{c|}{\multirow{3}{*}{Total}}  & \multicolumn{2}{c|}{FWD}                               & 12340.8          & 7423.95          & 1.662295678          & -        \\ \cline{3-8} 
\multicolumn{1}{|c|}{}                      & \multicolumn{1}{c|}{}                        & \multicolumn{2}{c|}{BWD}                               & 23360.5          & 14272.25         & 1.636777663          & -        \\ \cline{3-8} 
\multicolumn{1}{|c|}{}                      & \multicolumn{1}{c|}{}                        & \multicolumn{2}{c|}{Total}                             & 35701.3          & 21696.2          & 1.645509352          & -        \\ \hline
\multicolumn{2}{|c|}{\multirow{3}{*}{Others}}                                              & \multicolumn{2}{c|}{FWD}                               & 6874.3           & 7090.55          & -                    & -        \\ \cline{3-8} 
\multicolumn{2}{|c|}{}                                                                     & \multicolumn{2}{c|}{BWD}                               & 13920.7          & 14117.45         & -                    & -        \\ \cline{3-8} 
\multicolumn{2}{|c|}{}                                                                     & \multicolumn{2}{c|}{Total}                             & 20795            & 21208            & -                    & -        \\ \hline
\multicolumn{2}{|c|}{\multirow{3}{*}{Total}}                                               & \multicolumn{2}{c|}{FWD}                               & 19215.1          & 14514.5          & 1.323855455          & -        \\ \cline{3-8} 
\multicolumn{2}{|c|}{}                                                                     & \multicolumn{2}{c|}{BWD}                               & 37281.2          & 28389.7          & 1.313194574          & -        \\ \cline{3-8} 
\multicolumn{2}{|c|}{}                                                                     & \multicolumn{2}{c|}{\textbf{Total}}                    & \textbf{56496.3} & \textbf{42904.2} & \textbf{1.316801152} & -        \\ \hline
\multicolumn{4}{|c|}{\makecell{Masked\\decay}}                                                                                                                  & 0                & 45.2             & -                    & $\frac{1}{m}$      \\ \hline
\multicolumn{4}{|c|}{\makecell{Prune\\weights}}                                                                                                                  & 0                & 320.3            & -                    & $\frac{1}{m}$      \\ \hline
\multicolumn{4}{|c|}{\makecell{Transposable\\mask search}}                                                                                                    & 0                & 634.8            & -                    & $\frac{1}{40m}$ \\ \hline
\end{tabular}
\end{sc}
\end{small}
\end{center}
\vskip -0.1in
\end{table}

% \begin{center}
%   \begin{minipage}{\textwidth}
%     \setcounter{mpFootnoteValueSaver}{\value{footnote}} \centering
%      \begin{tabular}{l|l}
%        \textsc{Woman}             &\textsc{Relationship} \\ \hline 
%        Mona                       &Attached\footnotemark  \\ 
%        Diana Villiers             &Eventual wife  \\  
%        Christine Hatherleigh Wood &Fiance\footnotemark 
%      \end{tabular}
%   \end{minipage}%  percent sign keeps footnote text close to minipage
%   \stepcounter{mpFootnoteValueSaver}%
%     \footnotetext[\value{mpFootnoteValueSaver}]{%
%       Little is known other than her death.}%
%   \stepcounter{mpFootnoteValueSaver}%
%     \footnotetext[\value{mpFootnoteValueSaver}]{%
%       Relationship is unresolved in XXI.}
% \end{center}

%%%%%%%%%%%%%%%%%%%%%%%%%%%%%%%%%%%%%%%%%%%%%%%%%%%%%%%%%%%%%%%%%%%%%%%%%%%%%%%
%%%%%%%%%%%%%%%%%%%%%%%%%%%%%%%%%%%%%%%%%%%%%%%%%%%%%%%%%%%%%%%%%%%%%%%%%%%%%%%

\end{document}

%% file: common.tex
\usepackage{color}
\usepackage{bm}
\usepackage{hyperref}
\usepackage{amsmath,amsfonts,amsthm}
\usepackage{blindtext}
\usepackage{listings}
\usepackage{algorithm, algorithmic}
\usepackage{booktabs}
\usepackage{multirow}
\usepackage{multicol}
\usepackage{caption}
\usepackage{enumitem} % 加了这个包就work了，是否该加
\newcommand{\ie}{\emph{i.e.}}

\newlist{myitems}{enumerate}{3}
\setlist[myitems, 1]
{label=\arabic{myitemsi}.,leftmargin=15pt,labelwidth=10pt,labelsep=5pt,
topsep=0pt,parsep=0pt,partopsep=0pt,noitemsep
}

\usepackage{ifthen}

\newif\ifdebug
\debugfalse

\newcommand{\vect}[1]{\boldsymbol{\mathbf{#1}}}

\newcommand{\bv}{\vect b}
\newcommand{\cv}{\vect c}
\newcommand{\dv}{\vect d}

\newcommand{\gv}{\vect g}

\newcommand{\mv}{\vect m}

\newcommand{\uv}{\vect u}
\newcommand{\vv}{\vect v}
\newcommand{\wv}{\vect w}

\newcommand{\Av}{\vect A}
\newcommand{\Bv}{\vect B}
\newcommand{\Cv}{\vect C}

\newcommand{\Mv}{\vect M}
\newcommand{\Nv}{\vect N}

\newcommand{\Uv}{\vect U}
\newcommand{\Vv}{\vect V}
\newcommand{\Wv}{\vect W}
\newcommand{\Xv}{\vect X}

\newcommand{\Zv}{\vect Z}

\newcommand{\norm}[1]{\left\lVert#1\right\rVert}

\newcommand{\loss}{f_v}

\makeatletter
\newtheorem*{rep@theorem}{\rep@title}
\newcommand{\newreptheorem}[2]{%
	\newenvironment{rep#1}[1]{%
		\def\rep@title{#2 \ref{##1}}%
		\begin{rep@theorem}}%
		{\end{rep@theorem}}}
\makeatother